\documentclass[10pt,journal,compsoc]{IEEEtran}
\usepackage{amsmath,amsfonts}
\usepackage{algorithmic}
\usepackage{algorithm}
\usepackage{array}
\usepackage[caption=false,font=normalsize,labelfont=sf,textfont=sf]{subfig}
\usepackage{textcomp}
\usepackage{stfloats}
\usepackage{url}
\usepackage{verbatim}
\usepackage{graphicx}

\usepackage{multirow}  % 如果您使用多行单元格
\usepackage{longtable} % 如果表格太长
\usepackage[table,xcdraw]{xcolor}  % For coloring cells
\usepackage{graphicx}              % For resizing table
\usepackage{array}                 % For better column definition
\usepackage{caption}               % For better caption handling
\usepackage{tabularx}              % For auto-adjusting column width
\usepackage{booktabs}              % For Ablation study
\usepackage{diagbox}     
\usepackage{booktabs}  % 美观的表格
\usepackage{multirow}  % 合并单元格
\usepackage{bm}    %允许加粗对勾

\usepackage[most]{tcolorbox}  % 引入 tcolorbox 包

\begin{document}
\bibliographystyle{unsrt}
\title{PiercingEye: Dual-Space Video Violence Detection with Hyperbolic Vision-Language Guidance}

\author{Jiaxu~Leng, Zhanjie~Wu, Mingpi~Tan, Mengjingcheng~Mo, Jiankang~Zheng, Qingqing Li, Ji~Gan, and  \\Xinbo Gao, \IEEEmembership{Fellow, IEEE} 

\thanks{Jiaxu~Leng, Zhanjie~Wu, Mingpi~Tan, Mengjingcheng~Mo, Jiankang~Zheng, Ji~Gan, and Xinbo Gao are with the Key Laboratory of Image Cognition, Chongqing University of Posts and Telecommunications, Chongqing, China, and also with Guangyang Bay Laboratory, Chongqing Institute for Brain and Intelligence, Chongqing 400065, China.\\
Qingqing Li is with the State Key Laboratory of Intelligent Vehicel Safety Technology and China Automotive Engineering Research Institute Co., Ltd.}
\thanks{Corresponding author: Xinbo Gao.}

}

%\markboth{}%

%\IEEEpubid{}

\IEEEtitleabstractindextext{
\begin{abstract}

Existing weakly supervised video violence detection (VVD) methods primarily rely on Euclidean representation learning, which often struggles to distinguish visually similar yet semantically distinct events due to limited hierarchical modeling and insufficient ambiguous training samples. To address this challenge, we propose PiercingEye, a novel dual-space learning framework that synergizes Euclidean and hyperbolic geometries to enhance discriminative feature representation. Specifically, PiercingEye introduces a layer-sensitive hyperbolic aggregation strategy with hyperbolic Dirichlet energy constraints to progressively model event hierarchies, and a cross-space attention mechanism to facilitate complementary feature interactions between Euclidean and hyperbolic spaces. Furthermore, to mitigate the scarcity of ambiguous samples, we leverage large language models to generate logic-guided ambiguous event descriptions, enabling explicit supervision through a hyperbolic vision-language contrastive loss that prioritizes high-confusion samples via dynamic similarity-aware weighting. Extensive experiments on XD-Violence and UCF-Crime benchmarks demonstrate that PiercingEye achieves state-of-the-art performance, with particularly strong results on a newly curated ambiguous event subset, validating its superior capability in fine-grained violence detection.

\end{abstract}

}
\maketitle

\section{Introduction}
\label{introduction}

\IEEEPARstart{V}{ideo} Violence Detection (VVD) is a fundamental task in video understanding, focused on identifying violent or anomalous behaviors such as physical altercations, robberies, and traffic accidents in surveillance streams. Considering its vital applications in public security \cite{Sultani,Wu2020HLnet} and intelligent transportation \cite{yao2022dota}, VVD has become an increasingly important research area. Since violent events rarely occur in real-life situations, leading to the scarcity of annotations, several semi-supervised methods \cite{cao2024scene,cao2023new,leng2022anomaly,luo2021future} have been proposed to model the characterization patterns of normal events (non-violence) during the training phase and detect the outliers as violence during testing. However, these methods often struggle to effectively identify violence when encountering complicated or unseen scenes, leading to a high false alarm rate. Therefore, weakly-supervised methods \cite{Sultani,Wu2020HLnet,Wu2021LCTR} have received significant attention in recent years, where video-level labelled data is used to train the models. These methods handle VVD by a Multiple-Instance Learning (MIL) \cite{MIL}, in which a video is treated as a bag of instances (\textit{i.e. }snippets or segments), and their labels are predicted based on the video-level annotations.

\begin{figure*}[t]
    \centering
    \includegraphics[width=1\linewidth]{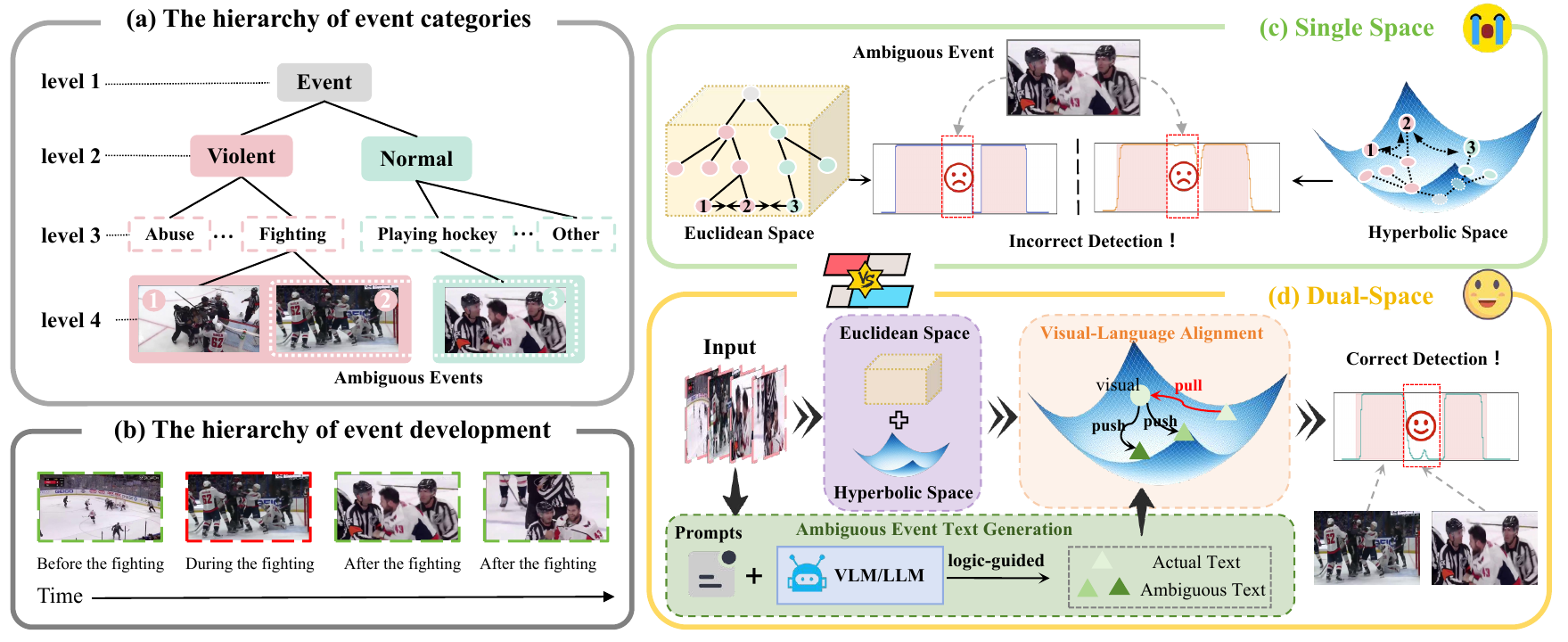}
   \caption{
Overview of the core idea behind the proposed PiercingEye framework.
(a) The hierarchical structure of event categories in VVD, where ambiguous events—such as fighting (violent) and playing hockey (normal)—are visually similar but semantically distinct, making them difficult to distinguish using conventional methods.
(b) The temporal hierarchy of event development, showing the semantic progression before, during, and after within an event, which provides contextual cues to mitigate ambiguity.
(c) Modeling in a single space—whether Euclidean or hyperbolic—struggles to simultaneously capture visual features and hierarchical event relations, often leading to incorrect predictions for ambiguous events.
(d) Our PiercingEye adopts a dual-space strategy: Euclidean space captures visual features, hyperbolic space models hierarchies, and logic-guided texts from VLMs/LLMs enable cross-modal alignment, enhancing ambiguous events detection.
}
    \label{fig:intro}
    \vspace{-14pt}
\end{figure*}

Based on the modal features of input data, existing weakly supervised VVD methods can be categorized into two primary paradigms: unimodal methods (relying solely on visual input) and multimodal methods (integrating both visual and auditory information). Unimodal methods \cite{Sultani,Tian2021RTFM,lV2023UMIL,zhou2023dual,pu2024learning,yang2024text,Feng2021mist,zhang2023exploiting,li2022MSL} distinguish violent behavior by extracting significant visual features, such as physical conflict and other motion features. 
However, these methods struggle with visually ambiguous events, where different semantic events share similar visual appearances. For instance, as shown in Fig. \ref{fig:intro}(a), normal body collisions and violent confrontations in ice hockey both involve rapid physical contact and clustered movements, making them difficult to distinguish using visual features alone.
While the former represents legal gameplay, the latter indicates aggression, highlighting the limitations of purely visual learning. 
To alleviate this, XD-Violence \cite{Wu2020HLnet} was introduced as a large-scale multimodal dataset that incorporates audio to enhance model discriminability. Although it has driven advances in multimodal VVD, most methods concentrate on either the problem of modality asynchrony \cite{Yu2022MACIL-SD,Leng2024MVFD} or the cross-modal fusion strategies \cite{Wu2022weakly,Pang2021violence,Pang2022AVD}, leaving the challenging yet crucial ambiguous violence events detection largely underexplored.
Moreover, even with multimodal input, existing VVD methods still encounter an insufficient representation learning at the feature level, which hampers the model's ability to make accurate distinctions between subtle differences in event semantics. 
This suggests that resolving ambiguity requires more than modality enhancement—it demands a deeper understanding of the event’s semantic structure and temporal dynamics. 
To fully understand an event, on the one hand, we need to explore the hierarchy of event categories, shown in Fig. \ref{fig:intro}(a). On the other hand, we need to sort through the events, including the trend before the event, the action during the event and the behaviour after the event, which reflects the hierarchy of event development, shown in Fig. \ref{fig:intro}(b). An ambiguous violent event is confusing at the current category level or happening moment, but may easily be detected if we can grasp the hierarchical relations. Fortunately, hyperbolic representation learning, characterized by exponentially increasing the metric distances and naturally reflecting the hierarchical structure of data, has gained attention and shown promising performance in computer vision tasks, like semantic segmentation \cite{atigh2022hyperbolic}, medical image recognition \cite{yu2022skin}, action recognition \cite{peng2020mix,long2020searching}, anomaly recognition \cite{hong2023curved}. At present, several methods \cite{HyperVD,ghadiya2024cross} have made preliminary attempts to introduce hyperbolic representation learning into the VVD task. HyperVD \cite{HyperVD} adopts Hyperbolic Graph Convolutional Network (HGCN) \cite{liu2019hyperbolic}, an extension of Euclidean graph convolution, to learn discriminative features in hyperbolic space. CFA-HLGAtt \cite{ghadiya2024cross} further incorporates a Lorentzian hyperbolic graph attention mechanism to capture spatial-temporal dependencies under weak supervision. However, both methods rely on rigid structural modeling strategies, such as hard-threshold node selection or static graph construction, which limit the effective modeling of hierarchical relations in hyperbolic space. 
In addition, existing VVD methods deploy feature embedding either in Euclidean or hyperbolic spaces. Representation learning in a single space is like picking the sesame and losing the watermelon, where the feature embedding is insufficient to guarantee the performance of VVD. On the one hand, hyperbolic representation learning strengthens the hierarchical relation of events but weakens the expression of visual features, on the other hand, Euclidean representations emphasize visual features but ignore relations between events. Therefore, leveraging the advantages of both spaces is essential for improving the performance of VVD methods, shown in Fig. \ref{fig:intro}(d).

Factually, a better approach to address ambiguous events may be collecting more relevant videos for explicit supervised learning based on dual-space modeling. However, it is challenging to obtain sufficient ambiguous samples, and video generation techniques fail to ensure the quality of generated videos. In recent years, we have witnessed the rapid development of large language models (LLMs) (e.g., GPT \cite{achiam2023gpt}, T5 \cite{raffel2020exploring}) and vision-language models (VLMs) (e.g., InternVL2 \cite{chen2024internvl}), which have demonstrated exceptional capabilities in understanding and generating human-like language. Therefore, a straightforward idea is to leverage ambiguous event texts generated by VLMs and LLMs as pseudo-supervision for ambiguous event videos. However, some challenges are coming: (1) How to generate textual descriptions that effectively capture the subtle differences of ambiguous events? (2) How to utilize the generated textual descriptions to enhance the model’s ability to distinguish ambiguous events?

To address these challenges, we analyze the nature of ambiguous events and how to express them effectively in language.
A crucial aspect of ambiguous events is that they exhibit high visual similarity while differing semantically. Therefore, constructing text descriptions that preserve this characteristic is essential for ensuring that they can serve as effective training signals. Our observation suggests that modifying either the scene or behavior of an event allows us to alter its semantic category while maintaining visual consistency. For instance, modifying the behavior while keeping the scene unchanged can transform ``fighting" into ``a normal body collision" both of which involve rapid body movements but have entirely different semantic meanings. Conversely, modifying the scene while keeping the behavior unchanged, such as changing a ``collision in a stadium" to a ``collision on the stree" results in a different semantic interpretation despite the same physical action. This approach enables the generation of text descriptions that are visually similar but semantically distinct, thus improving the model's ability to differentiate ambiguous events.
Based on this, we leverage large language models (LLMs) to automatically generate ambiguous event text descriptions following the aforementioned logic. This method not only systematically constructs data that aligns with the characteristics of ambiguous events but also enhances the model’s learning capability through textual information, allowing it to capture subtle semantic differences even when visual features are highly similar, ultimately improving its ability to recognize ambiguous events.
                                
In this paper, we propose a novel method--\textbf{PiercingEye} for weakly supervised VVD under the multimodal input setting. Specifically, we designed four customized modules, the Hyperbolic Energy-constrained Graph Convolutional Network module (HE-GCN), the Dual-Space Interaction module (DSI), the Ambiguous Event Text Generation module(AETG)  and the Hyperbolic Vision-Language Guided Loss (HVLGL). 
Instead of adopting the hard node selection strategy in HGCN, HE-GCN selects nodes for message aggregation by our introduced layer-sensitive hyperbolic association degrees, which are dynamic thresholds determined by the message aggregation degree at each layer. To better align with the characteristics of hyperbolic spaces, we introduce the hyperbolic Dirichlet energy to quantify the extent of message aggregation. Benefiting from the dynamic threshold, the layer-by-layer focused message passing strategy adopted by HE-GCN not only ensures the efficiency of information excavation but also improves the model's comprehensive understanding of the events, thus enhancing the model's ability to discriminate ambiguous violent events.  Although hyperbolic representation learning enhances the understanding of hierarchical relations of events, the role of visual representations in violence detection cannot be discarded. Subsequently, to break the information cocoon of two different geometric spaces, DSI utilises cross-space attention to facilitate information interactions between Euclidean and hyperbolic space to capture better discriminative features, where Euclidean representations have effectiveness on the significant motion and shape changes in the video, while hyperbolic representations accelerate the comprehension of hierarchical relations between events, working together to improve the performance of violence detection in videos. 
 
Furthermore, to effectively enhance the model’s ability to discriminate ambiguous events, AETG systematically modifies the scene or behavior of an event, utilizing the powerful language generation capabilities of vision-language models (VLMs) and large language models (LLMs) to generate ambiguous event texts. These generated texts are visually similar but semantically distinct, thus constructing representations of ambiguous events.
Finally, the HVLGL module addresses the issue of ineffective learning from hard ambiguous events due to the limitations of Euclidean space metrics and uniform negative sample treatment, by introducing the hyperbolic metric (Lorentzian metric) to better quantify the hierarchical semantic relationships and a dynamic weighting mechanism based on text similarity to prioritize optimization of more confusing ambiguous texts, thus improving the model's ability to distinguish ambiguous events.

At last, we remind that DSRL consisting of the HE-GCN and DSI modules was first introduced in our previous work \cite{leng2024beyond}. Compared to the preliminary version, in this paper, our contribution is multifold : 
\textbf{(1)} Building upon the implicit hierarchical modeling in our conference version, we address the scarcity of ambiguous video data by introducing ambiguous event texts, which explicitly enhance the model’s understanding of such events. This significantly improves the discrimination of ambiguous violence and achieves state-of-the-art performance on the XD-Violence dataset under both unimodal and multimodal settings. 
\textbf{(2)} To explicitly enhance the model's discrimination of ambiguous events, we introduce the Ambiguous Event Text Generation module (AETG), a systematic approach to generate textual descriptions for ambiguous events, which remains visually similar but semantically distinct. This method helps train the model to better recognize and differentiate ambiguous events by capturing the semantic differences that are otherwise difficult to detect with visual features alone.
\textbf{(3)} To enhance the model's ability to learn from ambiguous events, we propose the HVLGL module, which combines hyperbolic space with a dynamic weighting mechanism based on text similarity, ensuring that the model prioritizes learning from the most difficult ambiguous events and improving generalization.
\textbf{(4) }We provide a comprehensive survey of relevant works in the VVD field and further explore the application of large language models (LLMs) and vision-language models (VLMs) in VVD. 
\textbf{(5)} We present thorough quantitative analysis, including tests on a curated ambiguous event subset from the UCF-Crime dataset, validating the performance of our method in handling ambiguous events.
\textbf{(6)} To further validate our method, we conduct a series of visualizations, including the ambiguous text descriptions generated by AETG, comparisons of different HVLGL loss designs, feature visualizations for key modules, and ablation studies on critical parameters, all of which collectively demonstrate the effectiveness of our approach.

\section{Related Work}
\label{related_work}
\subsection{Video Violence Detection} 
Video violence detection (VVD) methods are generally categorized into hand-crafted feature-based and deep learning-based approaches. Early methods \cite{Gao2016Violence,mahmoodi2019classification,zhang2016discriminative,Mohammadi2016angry} primarily relied on hand-crafted visual features such as SIFT, STIP, HOG, HOF, and motion intensity, typically applied to small-scale datasets. Some audio-visual approaches  \cite{zhang2014mic,giannakopoulos2010audio,giannakopoulos2010multimodal} further incorporated audio features like MFCC, energy entropy, ZCR, and spectrogram to boost performance. However, the limited expressive capacity of manually designed features hindered their ability to capture complex and hierarchical patterns.
With the powerful learning capacity of deep neural networks, many deep learning-based methods have been proposed. However, the rarity of violent events in real-world data results in limited annotations. To mitigate this, some semi-supervised approaches \cite{cao2024scene,leng2022anomaly,cao2023new,luo2021future,luo2019video} model the distribution of normal patterns during training and detect deviations at test time, but often suffer from high false alarm rates in complex or unseen scenes.  Recently, weakly supervised methods have gained traction in VVD, using only video-level labels to train models without requiring frame-level annotations. 
These approaches can be divided into two main categories: unimodal (vision-based) and multimodal (audio-visual-based) methods. \\
\textbf{\textit{Unimodal methods.}} Unimodal VVD methods rely solely on visual cues to identify violent events.
Sultani et al. \cite{Sultani} first introduce an MIL-based method with a novel ranking loss, marking the beginning of a series of MIL-driven approaches. Since then, a large number of MIL-based methods have been proposed. 
Some works \cite{Tian2021RTFM,lV2023UMIL,zhou2023dual} focus on improving the MIL framework by optimizing snippet selection to better localize anomalies. Others \cite{pu2024learning,yang2024text} incorporate prompt learning into MIL, aiming to leverage semantic priors to enhance feature representation and detection performance. 
In addition, another line of research \cite{Feng2021mist,zhang2023exploiting,li2022MSL} explores two-stage self-training paradigms, where pseudo labels are first generated and then used to supervise a subsequent classification module, enabling more accurate and fine-grained anomaly scoring.
Unimodal methods enhance representation and discrimination by focusing on visual cues, but often neglect cross-modal interactions and complementary audio information, limiting detection accuracy and robustness. \\
\textbf{\textit{Multimodal methods.}} To alleviate this problem, Wu et al. \cite{Wu2020HLnet} released a large multimodal violence dataset, named XD-Violence, and accelerated a series of multimodal VVD methods \cite{Wu2020HLnet,Yu2022MACIL-SD,Leng2024MVFD,Wu2022weakly,Pang2021violence,Pang2022AVD}. In contrast to unimodal methods, they incorporate not only visual cues but also complementary audio information to improve the discrimination of ambiguous violent events. 
Subsequently, some studies \cite{Leng2024MVFD, Pang2022AVD} have focused on better integration of visual and audio information, and solving the modality asynchrony problem \cite{Yu2022MACIL-SD}.  Despite the progress of weakly supervised multimodal VVD methods,  all the above methods carry out representation learning in Euclidean Space, making it difficult to effectively handle ambiguous violence.  

\subsection{Hyperbolic Representation Learning}
Hyperbolic space, a non-Euclidean space with constant negative Gaussian curvature, is particularly effective at modeling hierarchical data structures due to its exponential volume growth with respect to radius, as opposed to the polynomial growth seen in Euclidean space. 
Initially, hyperbolic representation learning was primarily explored in natural language processing (NLP) \cite{nickel2017poincare,ganea2018hyperbolic,law2019lorentzian,shimizu2020hyperbolic,chen2021fully} and graph learning \cite{zhang2021hyperbolic, Chami2019hyperbolic,bachmann2020constant}. Owing to its strong capacity for modeling hierarchical relationships, it has recently attracted increasing attention in the computer vision community. In this domain, hyperbolic representation learning has demonstrated promising results across various tasks, including semantic segmentation \cite{atigh2022hyperbolic}, medical image classification \cite{yu2022skin}, action recognition \cite{peng2020mix, long2020searching}, and anomaly detection \cite{hong2023curved}.
More recently, HyperVD \cite{HyperVD} and CFA-HLGAtt \cite{ghadiya2024cross}
have made initial attempts at addressing the VVD task using hyperbolic representation learning. However, these methods focus solely on learning in hyperbolic space. While hyperbolic representation learning effectively captures the hierarchical relationships among events, it often compromises the fidelity of visual feature representation.
To address this limitation, we propose the first dual-space representation learning framework that leverages the complementary strengths of both Euclidean and hyperbolic spaces. 

\subsection{VLMs and LLMs for VVD}
In recent years, large language models (LLMs) have demonstrated remarkable capabilities in natural language generation, in-context learning, and knowledge reasoning \cite{achiam2023gpt,mann2020language,touvron2023llama,chiang2023vicuna}. Meanwhile, the rapid development of vision-language models (VLMs) \cite{chen2024internvl,radford2021CLIP,li2022blip} has endowed models with strong visual understanding and cross-modal reasoning abilities. These advances have significantly improved the effectiveness of tasks requiring both semantic comprehension and visual-linguistic interaction. In the field of video violence detection (VVD), several recent approaches have begun leveraging LLMs and VLMs to enhance model interpretability and generalization. For instance, some methods \cite{wu2024open,tang2024hawk,zanella2024harnessing} generate event descriptions based on user intent to facilitate weak supervision, while others \cite{zanella2024harnessing,ye2024vera,lv2024video} directly employ VLMs for visual representation learning and video understanding. 
Building upon this line of research, and aiming to balance model performance with computational cost, we adopt an open-source VLM to perform image understanding and generate initial descriptions. These descriptions are then further refined using a closed-source LLM, which produces semantically ambiguous yet visually plausible textual descriptions. \\
\textbf{\textit{Vision-Language alignment.}} 
Vision-language alignment losses are widely used in multimodal learning to align image or video features with textual semantics, typically through contrastive objectives. Representative methods such as CLIP \cite{radford2021CLIP} and BLIP \cite{li2022blip} learn a shared Euclidean embedding space where matched vision-language pairs are pulled together while mismatched pairs are pushed apart. This paradigm has shown strong performance in tasks like image-text retrieval \cite{li2022image} and Zero-Shot Semantic Segmentation \cite{li2024tagclip}. 
In VVD, recent works \cite{pu2024learning,yang2024text,wu2024open} have adopted vision-language alignment to improve representation learning, often leveraging CLIP for feature extraction. 
However, most of these approaches rely on static contrastive losses in Euclidean space, which are limited in capturing the hierarchical structure and semantic ambiguity inherent in complex events. In addition, negative samples are typically treated uniformly, ignoring their varying semantic difficulty—an issue that can be particularly problematic when aligning visually similar but semantically divergent actions (e.g., ``fighting'' vs.``a normal body collision'').
To address these limitations, we propose a hyperbolic vision-language alignment loss that leverages hierarchical geometry and adaptive weighting to better capture fine-grained visual-textual relationships under weak supervision.

\section{Preliminaries}
\label{preliminaries}
\textbf{Problem Definition.} Given a video sequence $S=\left \{ S_{t} \right \} _{t=1}^{T} $ with $T$ non-overlapping segments. For a video segment, the weakly supervised VVD requires distinguishing whether it contains violent events via an events relevance label $y_{t}\in \left \{ 0,1 \right \} $, where $y_{t}=1$ means in the current segment includes violent cues. \\
\textbf{Hyperbolic Geometry.} A Riemannian manifold $(\mathcal{M},g )$ of dimension \textit{n} is a real and smooth manifold equipped with an inner product on tangent space $g_{x}$: $\mathcal{T}_{x}\mathcal{M}  \times \mathcal{T}_{x}\mathcal{M}\to \mathbb{R} $ at each point $x\in \mathcal{M}$, where the tangent space $\mathcal{T}_{x}\mathcal{M} $ is a \textit{n}-dimensional vector space and can be seen as a first-order local approximation of $\mathcal{M}$ around point $x$. In particular, hyperbolic space $(\mathbb{D}_{c}^{n},g^{c} )$, a constant negative curvature Riemannian manifold, is defined by the manifold $\mathbb{D}_{c}^{n} = \left \{ x\in \mathbb{R}^{n}:c\left \| x \right \|< 1   \right \}$ equipped with the following Riemannian metric:$g_{x}^{c} = \lambda _{x}^{2}g^{E}$, where $\lambda _{x}:=\frac{2}{1-c\left \| x \right \|^{2} } $ and $g^{E}=\textit{I}_{n}$ is the Euclidean metric tensor. 
Considering the numerical stability and calculation simplicity of its exponential and logarithmic maps and distance functions, we select the Lorentz model \cite{Lorentz} as the framework cornerstone. \\
\textbf{Lorentz Model.} Formally, an $n$-dimensional Lorentz model is the Riemannian manifold $\mathbb{L}_{K}^{n} =\left ( \mathcal{L}^{n},\mathfrak{g}_{\textbf{x} }^{K}   \right )   $. $K$ is the constant negative curvature. $\mathfrak{g}_{\textbf{x} }^{K} = diag\left ( -1,1,\cdot \cdot \cdot ,1 \right ) $ is the Riemannian metric tensor. We denote $\mathcal{L}  ^{n}$ as the n-dimensional hyperboloid manifold with constant negative curvature $K$:
\begin{equation}
    \mathcal{L}  ^{n} : = \left \{ \textbf{x}\in \mathbb{R}^{n+1}: \left \langle \textbf{x},\textbf{x}   \right \rangle_{\mathcal{L} } = \frac{1}{K} ,x_{0}>0     \right \} .
\end{equation}
Let $\textbf{x} ,\textbf{y}  \in \mathbb{R} ^{n+1} $, then the {Lorentzian scalar product} is defined as: 
\begin{equation}
\label{Eq:lorentzian scalar product}
    \left \langle \textbf{x} ,\textbf{y} \right \rangle _{\mathcal{L} } : = -x_{0}  y_{0}+\sum_{i=1}^{n} x_{i}y_{i},
\end{equation}
where $\mathcal{L}  ^{n}$ is the upper sheet of hyperboloid in an (n+1)-dimensional Minkowski space with the origin $\left ( \sqrt{-1/K} ,0,\cdot \cdot \cdot,0  \right ) $. For simplicity, we denote point $x$ in the Lorentz model as $x\in \mathbb{L}_{K}^{n} $.\\
\textbf{Tangent Space.}  Given the tangent space at $x$ is defined as an n-dimensional vector space approximating $\mathbb{L}_{K}^{n}$ around $x$,
\begin{equation}
    \mathcal{T}_{\mathbf{x}} \mathbb{L}_{K}^{n}:=\left\{\mathbf{y} \in \mathbb{R}^{n+1} \mid\langle\mathbf{y}, \mathbf{x}\rangle_{\mathcal{L}}=0\right\}.
\end{equation}
Note that $\mathcal{T}_{\mathbf{x}} \mathbb{L}_{K}^{n}$ is a Euclidean subspace of $\mathbb{R} ^{n+1}$. Particularly, we denote the tangent space at the origin as $\mathcal{T}_{\mathbf{0}} \mathbb{L}_{K}^{n}$.\\
\textbf{Logarithmic and Exponential Maps.} 
The connections between hyperbolic space and tangent space are established by the exponential map $\exp _{\mathbf{x}}^{K}(\cdot)$ and logarithmic map $\log _{\mathbf{x}}^{K}(\cdot)$ are given as follows:
\begin{equation}
\label{Eq:Exp}
\begin{split}
\mathrm{exp}_{\mathbf{x}}^{K}(\mathbf{v}) =\ 
& \cosh\left(\sqrt{-K}\left\|\mathbf{v}\right\|_{\mathcal{L}}\right) \mathbf{x} \\
& + \sinh\left(\sqrt{-K}\left\|\mathbf{v}\right\|_{\mathcal{L}}\right) 
\frac{\mathbf{v}}{\sqrt{-K}\left\|\mathbf{v}\right\|_{\mathcal{L}}}
\end{split}
\end{equation}
\begin{equation}
    \log _{\textbf{x}}^{K}\left ( \textbf{y}  \right ) =d_{\mathbb{L} }^{K}\left ( \textbf{x},\textbf{y}\right )\frac{\textbf{y}-K\left \langle \textbf{x},\textbf{y}   \right \rangle_{\mathcal{L} }  }{\left \| \textbf{y}-K\left \langle \textbf{x},\textbf{y}   \right \rangle_{\mathcal{L} } \right \| }_{\mathcal{L} } ,
\end{equation}
where$\left \| \textbf{v}  \right \|_{\mathcal{L} }=\sqrt{\left \langle \textbf{v},\textbf{v}   \right \rangle_{\mathcal{L}}} $ denotes Lorentzian norm of \textbf{v} and $d_{\mathcal{L} }^{K}\left ( \cdot,\cdot \right )$ denotes  the Lorentzian intrinsic distance function between two points $\textbf{x},\textbf{y}\in \mathbb{L}_{K}^{n}$, which is given as:
\begin{equation}
    d_{\mathcal{L} }^{K}\left ( \textbf{x},\textbf{y}\right )=arcosh\left ( K\left \langle \textbf{x},\textbf{y}   \right \rangle_{\mathcal{L}} \right ) .
\end{equation}

\section{Methodology}
\label{method}
\begin{figure*}[ht]
    \centering
    \includegraphics[width=1\linewidth]{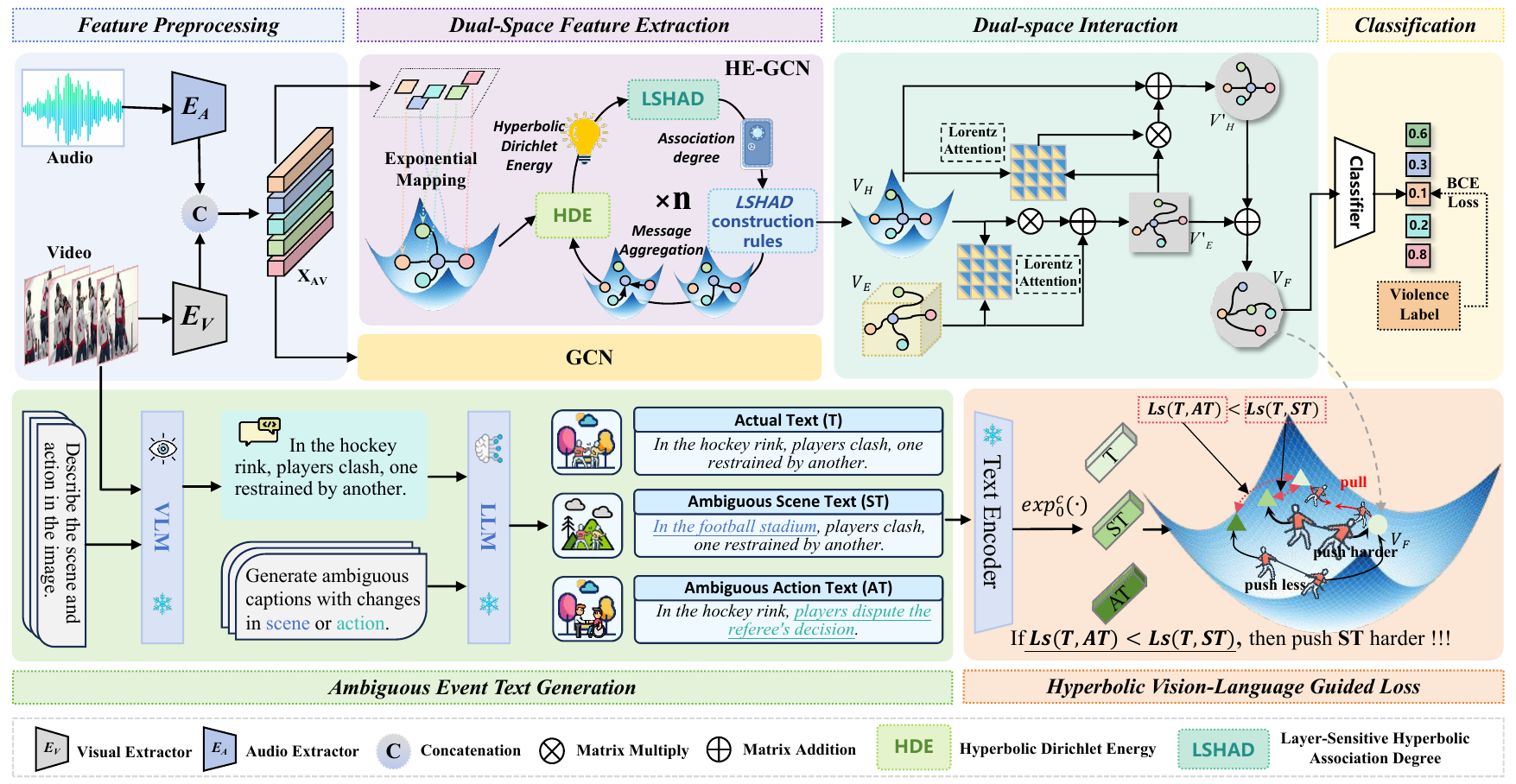}
    \caption{A conceptual diagram of our PiercingEye. After initial feature extraction by two encoders, visual features are learned using GCN in Euclidean space and hierarchical relationships are modeled through HE-GCN in hyperbolic space. These representations are then enhanced through interaction via the DSI module to improve feature discriminability. Meanwhile, 
    % VLM and LLM are used to generate ambiguous event descriptions, 
    the generated ambiguous event descriptions through VLM and LLM are applied with a novel hyperbolic vision-language guided loss to guide the model in learning more discriminative features. Finally, a classifier is used to obtain the violence prediction score.
    }
    \label{fig:framework}
    \vspace{-14pt}
\end{figure*}

In this paper, we propose the \textbf{PiercingEye} method to improve the discrimination of ambiguous violence. 
As depicted in Fig. \ref{fig:framework}, we first fuse multimodal features using two encoders and cross-modal attention to obtain initial representations, which are then fed into the dual-space feature extraction module, where parallel representation learning is performed. In Euclidean space, a Graph Convolutional Network (GCN) is used to learn the visual features of events, while the Hyperbolic Energy-constrained Graph Convolutional Network (HE-GCN) operates in hyperbolic space. HE-GCN captures the hierarchical context of events by constructing a layer-sensitive graph structure, enhancing the model's ability to understand both visual and contextual information.
The updated features from both spaces are then passed to the dual-space interaction module which employs cross-space attention, based on lorentzian metric, to enable information interaction between Euclidean and hyperbolic space, further enhancing feature discrimination. 
Next, in the ambiguous event text generation module, a VLM is used to describe video frames, and an LLM, guided by logical prompts, generates ambiguous event texts by altering scenes or actions. In the hyperbolic vision-language guided loss module, these textual features are encoded using the CLIP text encoder and mapped into hyperbolic space. Then, a contrastive learning loss with dynamic weighting—based on hyperbolic distance and text similarity—is applied to pull visual features closer to their corresponding text descriptions while effectively distinguishing them from semantically similar but ambiguous ones.
Finally, the refined features are sent to the classification module, which predicts the violence score for each video segment. 

\subsection{Feature Preprocessing}
\label{Feature_preprocess}
Following the previous works \cite{Wu2020HLnet,Pang2022AVD}, the visual and audio segments are processed by the I3D \cite{I3D} network pretrained on the Kinetics-400 dataset and the VGGish \cite{VGGish} network pretrained on a large YouTube dataset, respectively. After that, we perform further feature extraction using simple convolution and pooling operations. A simple cross-modal attention mechanism is then employed to enhance the audio features, which are subsequently concatenated with the visual features to form the fused features. 
We denote the resulting visual feature as ${{X}_{V}} \in \mathbb{R}^{T_{V  } \times d_{ V  } }  $ and the extracted audio feature is denoted as  $X_{A } \in \mathbb{R}^{T_{A  } \times d_{ A  } } $. For the rest of the paper, $T_{(\cdot )} $ and $d_{(\cdot )} $ are used to represent sequence length and feature dimension, respectively. Finally, the features from both modalities are concatenated to obtain the initial features as $X \in \mathbb{R}^{T \times (d_{ A  } +d_{V})} $. 

\subsection{Dual-Space Representation Learning} 
\label{Dual_Space Representation Learning}
Existing VVD methods deploy feature embedding either in Euclidean or hyperbolic spaces. The feature embedding in a single space is insufficient to guarantee the performance of VVD methods. To this end, we introduce a dual-Space representation learning paradigm, which consists of two key modules: the Dual-Space Feature Extraction (DSFE), where representation learning is performed separately in both spaces and the Dual-Space Interaction (DSI), which employs cross-space attention to facilitate interactions between Euclidean and hyperbolic spaces. 
Specifically, in the DSFE, to enhance hierarchical semantic modeling, we introduce HE-GCN for hyperbolic representation learning, enabling better capture of hierarchical relationships. Meanwhile, GCN is employed in Euclidean space for feature learning, complementing the visual representation.
\subsubsection{Hyperbolic Energy-constrained Graph Convolutional Network Module (HE-GCN)}
\label{HE-GCN} 
\begin{figure}[t]
    \centering
    \includegraphics[width=1\linewidth]{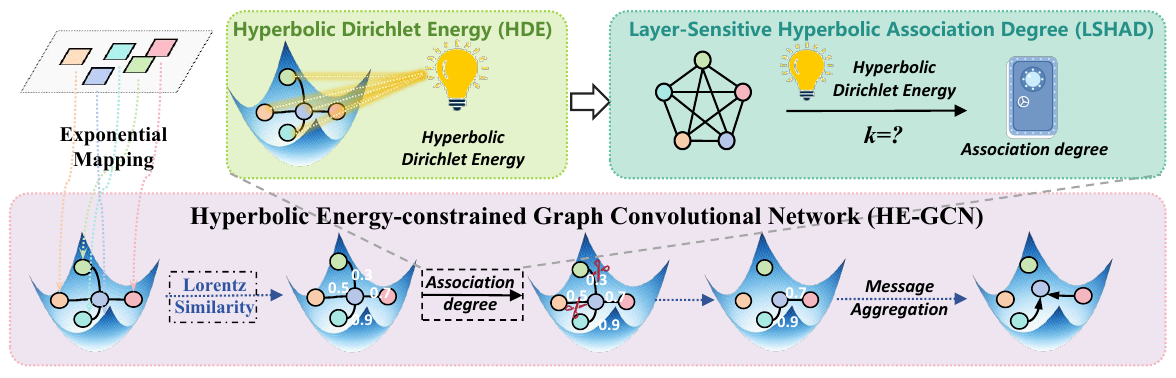}
    \caption{An illustration of the proposed HE-GCN. We first exponentiate the features into hyperbolic space and compute the lorentz similarity between nodes. Then, we calculate the hyperbolic Dirichlet energy and the layer-sensitive hyperbolic association degrees, which are used to construct the message graph, followed by message aggregation.
    }
    \label{fig:HEGCN}
    \vspace{-14pt}
\end{figure}

As depicted in Fig. \ref{fig:HEGCN}, HE-GCN primarily involves mapping features from Euclidean space to hyperbolic space, then transforming the features, constructing the message graph by calculating hyperbolic Dirichlet energy and Layer-Sensitive Hyperbolic Association Degree, and finally aggregating messages to obtain the new feature graph. \\
\textbf{Mapping from Euclidean to hyperbolic spaces.}
Let $\left \{ x_{i}^{E} \right \}_{i\in \mathcal{V} } $ be input Euclidean node features, and $\textbf{o}: =[1,0,\cdot \cdot \cdot ,0]$ denote the origin on the manifold $\mathcal{L} $ of the Lorenzt model. 
There is $\left \langle \textbf{o}, \left [0,x_{i}^{E}  \right ]\right \rangle _{\mathcal{L} }=0$, where $\left \langle \cdot ,\cdot \right \rangle _{\mathcal{L} }$ denotes the Lorentz inner product defined in Eq. \ref{Eq:lorentzian scalar product}. We can reasonably regard $\left [0,x_{i}^{E}  \right ]$ as a node on the tangent space at the origin \textbf{o}. 
HE-GCN uses the exponential map defined in Eq. \ref{Eq:Exp} to generate hyperbolic node representations on the Lorentz model: 
\begin{equation}
     \begin{aligned}
    x_{i}^{\mathcal{L}} &= \mathrm{exp}_{\textbf{o}}\left( \left[ 0, x_{i}^{E} \right] \right) \\
    &= \left[ \cosh\left( \left\| x_{i}^{E} \right\|_{2} \right), \sinh\left( \left\| x_{i}^{E} \right\|_{2} \right) \frac{x_{i}^{E}}{\left\| x_{i}^{E} \right\|_{2}} \right].
     \end{aligned}
\end{equation}
\textbf{Hyperbolic Feature Transformation.} According to \cite{chen2021fully}, we  reformalize the lorentz linear layer to learn a matrix $\textbf{M} =\begin{bmatrix}
 \textbf{v}^{\top} \\ \textbf{W} 

\end{bmatrix}$, $\textbf{v} \in \mathbb{R}^{n+1}$, $\textbf{W}\in \mathbb{R}^{m\times(n+1)}$ satisfying $\forall \textbf{x} \in \mathbb{L}^{n}  $, $f_{\textbf{x}}(\textbf{M})\textbf{x} \in \mathbb{L}^{m} $, where $f_{\textbf{x} }: \mathbb{R}^{(m+1)\times (n+1)}\to  \mathbb{R}^{(m+1)\times (n+1)}$ should be a function that maps any matrix to a suitable one for the hyperbolic linear layer. Specifically, $\forall \textbf{x} \in \mathbb{L}_{K}^{n} $, $\textbf{M}\in \mathbb{R}^{(m+1)\times(n+1)}$, $f_{\textbf{x}}(\textbf{M})$ is given as:
\begin{equation}
    f_{\textbf{x}}(\textbf{M})=f_{\textbf{x} }\left ( \begin{bmatrix}
 \textbf{v}^{\top}  \\ \textbf{W} 

\end{bmatrix} \right ) =\begin{bmatrix}
 \frac{\sqrt{\left \|  W\textbf{x}  \right \|^{2}-1/K }}{\textbf{v}^{\top}\textbf{x}  }\textbf{v}^{\top} \\
\textbf{W} 

\end{bmatrix}.
\end{equation}
\newtheorem{theorem}{Theorem}
\begin{theorem}
\label{thm:1}
$\forall \textbf{x} \in \mathbb{L}^{n}  $, $\textbf{M}\in \mathbb{R}^{(m+1)\times(n+1)}$, we have $f_{x}(\textbf{M})\textbf{x} \in \mathbb{L}_{K}^{m}$.
\end{theorem}

For simplicity, we use a general formula $*$ of  hyperbolic linear layer for feature transformation based on $f_{\textbf{x} }\left ( \begin{bmatrix}
 \textbf{v}^{\top}  \\ \textbf{W} 

\end{bmatrix} \right ) \textbf{x} $ with activation, dropout, bias and normalization,
\begin{equation}
    \textbf{y}=HL\left ( \textbf{x}  \right ) =\begin{bmatrix}
 \sqrt{\left \| \phi \left (\textbf{W}\textbf{x},\textbf{v}  \right )  \right \|^{2}-1/K } \\ \phi \left (\textbf{W}\textbf{x},\textbf{v}  \right ) 

\end{bmatrix},
\end{equation}
where $\textbf{x} \in \mathbb{L}_{K}^{n}$, $\textbf{v} \in \mathbb{R}^{n+1}$, $\textbf{W} \in \mathbb{R}^{m\times \left ( n+1 \right ) }$, and $\phi$ is an operation function: for the dropout, the function is $\phi \left ( \textbf{W}\textbf{x},\textbf{v}  \right ) = \textbf{W}dropout\left( \textbf{x}\right)$; for the activation and normalization $\phi \left ( \textbf{W}\textbf{x},\textbf{v}  \right ) = \frac{\lambda\sigma \left (  \textbf{v}^{\top}\textbf{x}+  b^{\prime}   \right )  }{\left \| \textbf{W}h\left ( \textbf{x}  \right )+\textbf{b}    \right \| }\left ( \textbf{W}h\left ( \textbf{x}  \right )+\textbf{b} \right )  $, where $\sigma$ is the sigmoid function, $\textbf{b}$ and $b^{\prime}$ are bias terms, $\lambda > 0$ controls the scaling range, $h$ is the activation function. And then, we need to construct the message graph. \\
\textbf{Hyperbolic Dirichlet Energy. }Given the hyperbolic embedings $\textbf{x} = \left \{\textbf{x}_{i} \in \mathbb{L}_{K} ^{d}    \right \} _{i=1}^{\left | \mathcal{V}  \right | }  $, the hyperbolic Dirichlet energy (\textit{HDE}) $E_{H}^{K }\left ( \textbf{x}  \right )  $ is defined as: 
\begin{multline}
    E_{H}^{K }\left ( \textbf{x}  \right )  = \\ 
    \frac{1}{2} \sum_{i,j=1} ^{N}d_{\mathcal{L} }^{K }\left ( \mathrm {exp}_{\textbf{o} }^{K }\frac{log_{\textbf{o} }^{K}(\textbf{x}_{i} )}{\sqrt{1+d_{i}} },\mathrm {exp}_{\textbf{o} }^{K } \frac{log_{\textbf{o} }^{K}(\textbf{x}_{j} )}{\sqrt{1+d_{j}} } \right )^{2} ,
\end{multline}
where $d_{i/j}$ denotes the node degree of node $i/j$. The distance $d_{\mathcal{L} }^{K}\left ( \textbf{x},\textbf{y}\right )$ between two points $\textbf{x},\textbf{y}\in \mathbb{L}$ is the geodesic.
Given that each node is connected to every other node in videos, resulting in a node degree \(d_i\) of \(n-1\) (where \(n\) is the total number of nodes), the formula for hyperbolic Dirichlet energy can be simplified:
\begin{equation}
    E_{H}^{K }\left ( \textbf{x}  \right )  =\frac{1}{2} \sum_{i,j=1} ^{N}d_{\mathcal{L} }^{K }\left ( \textbf{x}_{i},\textbf{x}_{j}  \right )^{2}  .
\end{equation}
\textit{HDE} is used to measure the similarity between node features in order to gauge the degree of information aggregation among features. It is evident that as hyperbolic message aggregation progresses, the similarity between features gradually decreases. Hyperbolic message aggregation reduces \textit{HDE}. It can be expressed as: $E_{H}^{K }\left ( \textbf{x} ^{(l+1)} \right )\le E_{H}^{K }\left ( \textbf{x} ^{(l)} \right )$, where $l$ is the layer number. \\
\textbf{Layer-Sensitive Hyperbolic Association Degree.} Based on \textit{HDE}, we design \textit{Layer-Sensitive Hyperbolic Association Degree (LSHAD)} to guide the node selection strategy for our construction of graphs for message aggregation. It is defined as: 
\begin{equation}
\begin{aligned}
    LSHAD_k &= f\left ( E_{H}^{K}(x),k \right ) \\
            &= \text{sigmoid}\left(\beta k - \gamma + \frac{1}{E_{H}^{K}(x)+1} \right),
\end{aligned}
\end{equation}
where $f\left ( \cdot  \right )$ is a function related to $k$ and $E_{H}^{K}(x)$, $k$ is the current layer number. $\beta$  and $\gamma$ are hyperparameters. 
\textbf{Lorentzian similarity.} Based on the Lorentzian distance, the Lorentzian similarity to measure the feature semantic similarity between nodes is given by
\begin{equation}
\label{ls}
    Ls\left ( x_{i},x_{j} \right ) = exp(-d_{\mathcal{L}}^K(x_{i},x_{j})),
\end{equation}
 where $d_{\mathcal{L}}^K(\cdot,\cdot)$ is the Lorentzian intrinsic distance function. We define the initial adjacent matrix $A^{\mathbb{L}}\in \mathbb{R}^{T \times T}$ via lorentz similarity:
\begin{equation}                A_{i,j}^{\mathbb{L}}=softmax(Ls(x_{i},x_{j})).
 \end{equation}
\textbf{LSHAD Construct rules.} With \textit{LSHAD}, we propose message graph construction rules, called \textit{LSHAD} construct rules. It is defined as: 
\begin{equation}
    \begin{cases}
A^{\prime }_{i,j}=A_{i,j} ,& \text{ if } A_{i,j}\ge LSHAD^{k} \\
  A^{\prime }_{i,j}=0 ,& \text{ if } A_{i,j}<LSHAD^{k},
\end{cases}
\end{equation}
where $A_{i,j}$ means the lorentz similarity between nodes $i$ and $j$ in the graph $G$. Formally, it enforces the elements of the adjacency matrix $A$ that are less than \textit{LSHAD} to be zeros. Finally, we can dynamically construct message graphs at each layer via \textit{LSHAD} construct rules to obtain better contextual information.
Following \textit{LSHAD} construction rules, we can dynamically construct message graphs  $G^{\prime}$, which we then use to perform message aggregation operations to obtain better contextual information. \\
\textbf{Hyperbolic Message Aggregation. }We use graph $G^{\prime}$ to perform message aggregation, and the message aggregation can be defined as: 
\begin{equation}
    MA\left (\textbf{y}_{i} \right )=\frac{ {\textstyle \sum_{j=1}^{m}}A_{ij}\textbf{y}_{j}  }{\sqrt{-K}\left | \left \|  {\textstyle \sum_{k=1}^{m}}A_{ik}\textbf{y}_{k}   \right \|_{\mathcal{L}} \right |  } ,
\end{equation}
where $m$ is the number of nodes. $\mathrm {y}_{i} $ is the node features. 
\textbf{Temporal Context Aggregation. }In addition to semantic context, temporal relationships are crucial for video-based tasks. To capture temporal relation, we construct a temporal relation graph based on the video’s temporal structure. The adjacency matrix $A^{\mathbb{T}}$ for the temporal relation graph depends on the temporal positions of snippets, and we compute the temporal similarity using a temporal distance function: \begin{equation} A_{ij}^{\mathbb{T}} = \exp\left(-\frac{|i - j|}{\eta}\right), \end{equation}where $\eta$ controls the range of temporal influence.
We then use HE-GCN for semantic message aggregation and HGCN for temporal message aggregation. The results are concatenated to obtain a unified representation in hyperbolic space. 
\subsubsection{Dual-Space Interaction (DSI)} 
\label{Dual_Space Interaction}
Although hyperbolic representation learning enhances understanding of event hierarchies, visual representations remain crucial in violence detection. Fusing representations from different spaces is challenging; thus, DSI employs cross-space attention to facilitate interactions between Euclidean and hyperbolic spaces. \\
\textbf{Cross-Space Attention Mechanism.} 
Cross-Space Attention Mechanism utilizes the Lorentzian metric to calculate attention scores between nodes from different spaces, accurately measuring semantic similarity and better preserving their true relationships by computing the nonlinear distance between them. We denote the features in Euclidean space as $V_{E}$ and the features in hyperbolic space as $V_{H}$. $CSA_{E\to H}$ models the between-graph interaction and guides the transfer of inter-graph message from $V_E$ to $V_H$. 
First, we use linear layer to transform $V_H$ to the \textit{key} graph $V_k$ and the \textit{value} graph $V_v$, and $V_{E}$ to the \textit{query} graph $V_q$. Then, we use lorentzian metric to calculate the attention map $\mathcal{A}_{E\to H}$ as follows: 
\begin{equation}
    \begin{aligned}
    \mathcal{A}_{E\to H} &= softmax(Ls(V_{q},V_{k})), \\
    Ls(x_i,x_j) &= 
    \begin{cases}
        Ls(x_i,x_j), & \text{if } Ls(x_i,x_j) > \lambda,  \\
        0, & \text{if } Ls(x_i,x_j) \leq \lambda ,
    \end{cases}
    \end{aligned}
    \label{EGA}
\end{equation}
where $Ls(\cdot)$ is Lorentzian similarity defined in Eq. \ref{ls} and $\lambda$ is the threshold value to eliminate weak relations and strengthen correlations of more similar pairs.
The representation from $E$ to $H$ can be formulated as follows:
\begin{equation}
    \begin{aligned}
    V^{\prime}_{H} &= CSA_{E \to H}(V_{H},V_{E}) \\
    &= softmax\left(\frac{\mathcal{A}_{E\to H} \times V_{k}}{\sqrt{d}} \right) V_{v}.
    \end{aligned}
\end{equation}

The interaction process in DSI can be represented as follows:
\begin{equation}
    \begin{aligned}
        V_{E}^{\prime } &= \alpha \times CSA_{E\to H}(V_{E},V_{H})+ V_{E}, \\
V_{H}^{\prime } &= \alpha \times CSA_{H\to E}(V_{H},V_{E}^{\prime})+V_{H}, \\
V_{F} &= MaxPool([V_{E}^{\prime}\oplus V_{H}^{\prime} ]) ,
    \end{aligned}
    \label{CSA}
\end{equation}
where $ V_{E}^{\prime }$ represents the features obtained by enhancing Euclidean space features using hyperbolic space features, and $\alpha$ serves as a scaling factor that controls the contribution of the hyperbolic space features ($V_H$) to the enhanced Euclidean space features ($V_E^\prime$). $MaxPool$ is the max pooling operation and $\oplus$ represents the concatenation operation.\\
\begin{figure*}[t]
    \centering
    \includegraphics[width=0.9\linewidth]{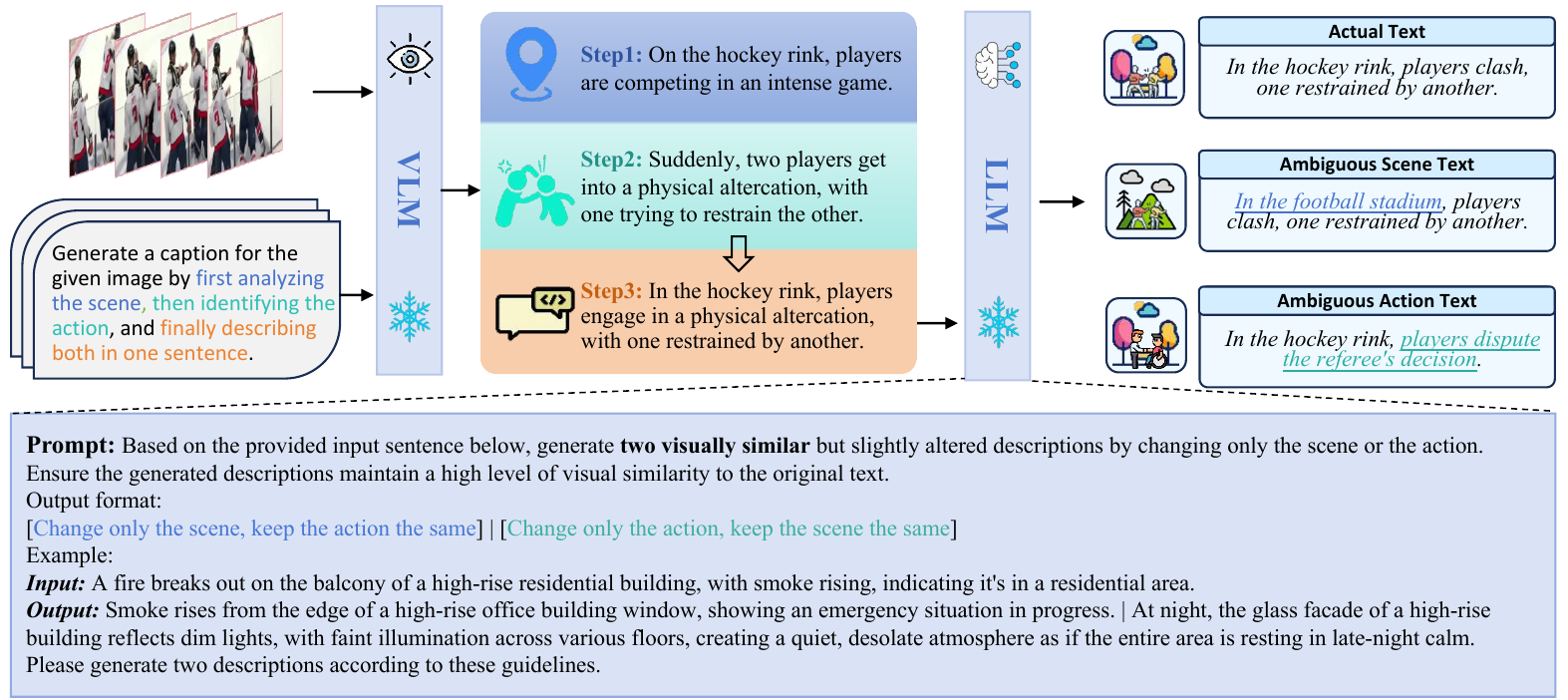}
    \caption{A conceptual diagram of our AETG. We first use a``scene analysis followed by behavior analysi'' approach to prompt the VLM to generate textual descriptions for each frame. Then, based on our designed scene-action reasoning, we guide the LLM to systematically generate ambiguous text descriptions from the previously generated ones. The visualization results of AETG can be found in Section \ref{fig:AETG_text_vis}.
    }
    \label{fig:AETG_framework}
    \vspace{-14pt}
\end{figure*}
\subsection{Ambiguous Event Text Generation (AETG)}
\label{Ambiguous Event Text Generation}
To explicitly enhance the model's ability by generating ambiguous event texts, AETG systematically modifies the scene or behavior of an event through the powerful language generation capabilities of VLMs and LLMs. This generates textual descriptions that are visually similar but semantically different, thereby constructing representations of ambiguous events. The illustration of our proposed AETG is shown in Fig. \ref{fig:AETG_framework}.
\subsubsection{Video Description Generation} 
To obtain natural language descriptions, we use a Vision-Language Model (VLM) to automatically generate captions. We first sample one frame every 16 frames to reduce the workload, which also aligns with the training process.  Then, we design prompts to instruct the VLMs (e.g.,  InternVL2-8B \cite{chen2024internvl}, BLIP2 \cite{li2023blip}) to first understand the scene, then the action, and ultimately describe the event in a single sentence, in order to better understand the events in the image.
This process can be representated as follows:
\begin{equation}
    T = \Phi _{VLM}(F,Prompt_{VLM}),
\end{equation}
where $T$ is the generated text description, $\Phi _{VLM}$ represents the VLMs, $F$ denotes the input video frames, and $Prompt_{VLM}$ refers to the prompt for the VLM. 
\begin{tcolorbox}[colback=green!5, colframe=gray!40, coltitle=black, title=$Prompt_{VLM}$]
    Generate concise captions based on the primary content of the input image. Each input includes a description of the scene and action, and a single sentence caption should be generated accordingly. If an anomaly is detected, the caption should focus solely on describing the scene and action related to the anomaly. If no anomaly is detected, generate a normal scene \& action description.
\end{tcolorbox}
\subsubsection{Ambiguous Description Generation}
For the generated text descriptions, we systematically modify the scene or action to obtain descriptions of ambiguous events. At the same time, we ensure that the generated descriptions remain visually similar to the given text descriptions, so as to better align with the characteristics of ambiguous events. The process can be representated as follows:
\begin{equation}
    [ST|AT] = \Phi _{LLM}(T,Prompt_{LLM}),
\end{equation}
where \( ST \) refers to the text description generated by modifying the scene, while \( AT \) refers to the text description produced by modifying the action. \( \Phi_{LLM} \) represents the LLMs, and \( Prompt_{LLM} \) refers to the prompt provided to the LLM. \( Prompt_{LLM} \) is shown in Fig. \ref{fig:AETG_framework}

\subsection{Hyperbolic Vision-Language Guided Loss (HVLGL)}
\label{Hyperbolic V-L guided loss}
While generating ambiguous event texts provides valuable training signals, effectively integrating them into the model remains a challenge. Traditional contrastive learning methods struggle to capture the hierarchical nature of textual semantics and often treat all negative samples equally, ignoring their varying levels of difficulty. To address these issues, we propose the Hyperbolic Vision-Language Guided Loss (HVLGL), which leverages hyperbolic space representation and a dynamic weighting mechanism to improve feature alignment and enhance the model’s ability to distinguish ambiguous events.

Given a batch of visual features $V$ and text features $T$, we first encode the text representations using a pre-trained text encoder (e.g., CLIP). To better capture hierarchical semantics, we project the encoded text features into hyperbolic space using the exponential map:
\begin{equation}
    T^{\mathcal{L}} = \exp_{\mathcal{L}} \left( \text{TextEncoder}(T) \right),
\end{equation}
where $\text{TextEncoder}(\cdot)$ denotes the text encoding function. $\exp_{\mathcal{L}}(\cdot)$ is the exponential map that projects features from Euclidean space to the Lorentz manifold $\mathcal{L}$. $T^{\mathcal{L}}$ represents the resulting text features embedded in hyperbolic space. We compute the positive and negative sample similarities as follows:
\begin{equation}
\begin{split}
    S^+ &= Ls(V, T^{+,\mathcal{L}}), \\
    S^-_j &= Ls(V, T^{-,\mathcal{L}}_j), \quad j = 1, \dots, N.
\end{split}
\end{equation}
where $Ls(\cdot)$ is Lorentzian similarity defined in Eq. \ref{ls}. 

To prioritize harder negative samples, we introduce a text similarity-based dynamic weighting mechanism:
\begin{equation}
    w_j = \exp(-\theta  Ls(T^{+,\mathcal{L}}, T^{-,\mathcal{L}}_j)),
\end{equation}
where $\exp(\cdot)$ is the natural exponential function, $\theta $ is a scaling hyperparameter,  $T^{+,\mathcal{L}}$ and $T^{-,\mathcal{L}}_j$ denote the positive and $j$-th negative text features embedded in hyperbolic space, respectively. $w_j$ is the weight for the $j$-th negative sample, where smaller distances yield higher weights, emphasizing harder negatives and guiding the model to focus on more challenging distinctions. The weighted negative similarities are then computed as:
\begin{equation}
    S^-_j = w_j Ls(V, T^{-,\mathcal{L}}_j).
\end{equation}

The final HVLGL loss is defined as:
\begin{equation}
    \mathcal{L}_{\text{HVLGL}} = -\mathbb{E} \left[ \log \frac{\exp(S^+ / \tau)}{\exp(S^+ / \tau) + \sum\limits_{j=1}^{N} \exp(S^-_j / \tau)} \right].
\end{equation}
where $\tau$ is the temperature parameter. This loss encourages the model to align visual features with positive textual representations while dynamically adjusting the optimization focus towards harder ambiguous samples, improving its ability to distinguish ambiguous events.
   
\subsection{Classification}
\label{Classification}
\textbf{Classifier.} 
As shown in Fig. \ref{fig:framework}, we input the enhanced embeddings $V_{F}$ from DSI into hyperbolic classifier utilizing Lorentzian metric, which can be formalized as: 
\begin{equation}
    S=\sigma (\epsilon +\epsilon \left \langle V_{F},W \right \rangle_{\mathcal{L} }+b ),
\end{equation}
where $\sigma$ is sigmoid function and $W$ is weight matrices. $b$ and $\epsilon$ denotes bias term and hyper-parameter, respectively. Lastly, we supervise the training of the violence scores obtained by the model with the real labels using the binary cross-entropy loss. \\
\textbf{Objective Function.} We use binary cross-entropy as our classification loss. Its calculation formula is:
\begin{equation}
    L_{CLS}=-\frac{1}{N} \sum_{i=1}^{N} \left ( y_{i}\log(\hat{y} _{i}) +(1-y_{i})log(1-\hat{y} _{i})  \right ) ,
\end{equation}
where $y_{i}$ is true label, $\hat{y}_{i}$ is the predicted label, $N$ is the batch size.  

To jointly optimize the representation learning and classification objectives, we define the overall loss function as a combination of the HVLGL and the classification loss. The final loss function is formulated as:
\begin{equation}
    \mathcal{L} = \psi  \mathcal{L}_{\text{HVLGL}}(V_{F},T) + \mathcal{L}_{\text{CLS}},
\end{equation}
where $\mathcal{L}_{\text{HVLGL}}$ is the hyperbolic vision-language guided loss that enhances representation learning by aligning visual features with textual descriptions while distinguishing ambiguous negative samples, and $\mathcal{L}_{\text{}}$ is the classification loss that supervises the final violence score prediction. $\psi $ is hyperparameter that balance the contributions of the two loss terms.

\section{Experiments}
\label{experiment}
\subsection{Experimental Setup}
\textbf{Datasets.} Under the multimodal input setting, we follow \cite{Wu2020HLnet,Yu2022MACIL-SD,HyperVD} to conduct experiments on XD-Violence, which is the only and extremely challenging VVD dataset with multimodal information. Under the unimodal input setting, both the XD-Violence and UCF-Crime datasets are used to evaluate our method. We conducted experiments on XD-Violence with both multi-modal input settings and single-modal input settings. Additionally, we performed experiments on UCF-Crime with single-modal input settings to demonstrate the generalization capability of PiercingEye. \textit{(1) XD-Violence dataset} \cite{Wu2020HLnet} is by far the only available large-scale audio-visual dataset for violence detection, which is also the largest dataset compared with other unimodal datasets. XD-Violence consists of 4,757 untrimmed videos (217 hours) and six types of violent events, which are curated from real-life movies and in-the-wild scenes on YouTube. For XD-Violence dataset, only video-level annotations are provided.
\textit{(2) UCF-Crime dataset }\cite{Sultani} is a large-scale dataset comprised of real-world videos captured by surveillance cameras. It consists of 1,610 training videos annotated with video-level labels and 290 test videos annotated at the frame level to facilitate performance evaluation. The videos are collected from different scenes and encompass 13 distinct categories of anomalies. \\
\textbf{Evaluation Metrics.} To quantitatively evaluate the performance, we follow standard pratice \cite{Wu2020HLnet,lV2023UMIL,Yu2022MACIL-SD}. For XD-Violence, we utilize the frame-level average precision (AP) as the evaluation metric. For UCF-Crime, we adopt the area under the curve of the frame-level receiver operating characteristic (AUC) to evaluate performance.\\
\textbf{Implementation Details.}  Follow the common setting for fair comparison, the visual sample rate is set to 24 fps, and visual features are extracted using a sliding window of 16 frames. For audio inputs, each audio track is divided into 960-ms overlapping segments, from which 96 $\times$ 64 log-mel spectrograms are computed. 
On UCF-Crime dataset, the visual features are extracted using a pre-trained CLIP \cite{radford2021CLIP}(ViT-B/16) visual encoder. The model is trained using the Adam optimizer \cite{Adam} (without weight decay) for 20 epochs with a batch size of 256 on the XD-Violence dataset, and AdamW \cite{loshchilov2019adamw} with a weight decay of 0.00005 for 10 epochs with a batch size of 64 on UCF-Crime. The initial learning rate is set to 0.001 for XD-Violence and 0.0005 for UCF-Crime. For learning rate scheduling, we adopt a cosine annealing scheduler \cite{Cosinescheduler} for XD-Violence, and a multi-step scheduler for UCF-Crime, which decays the learning rate by a factor of 0.1 at the 4th and 8th epochs.
Following \cite{HyperVD}, the temperature $\eta$ is empirically set to $e$. For the MIL-based learning, we set the value of $k$ in the $k$-max activation as $\left \lfloor \frac{T}{16}+1 \right \rfloor$, where $T$ denotes the temporal length of the input features.  \\
\begin{table}[t]
    \centering
    \caption{Comparison of frame-level AP performance on the XD-Violence dataset under unimodal and multimodal input settings.``*" refers to our previous work.
}
    \resizebox{\linewidth}{!}{
    \renewcommand{\arraystretch}{1.2}  
    \rowcolors{2}{gray!10}{white}  
    \begin{tabular}{l l c c c}  
        \toprule
        \rowcolor{black!10} \textbf{Methods} & \textbf{Venue} & \textbf{Input Setting} & \textbf{Feature Space} & \textbf{AP(\%)}\\
        \midrule
        Sultani et al.\cite{Sultani}& CVPR18 & Uni & E & 73.20 \\
        CRFD \cite{Wu2021LCTR}           & TIP21  & Uni & E & 75.90 \\
        RTFM \cite{Tian2021RTFM}           & ICCV21 & Uni & E & 77.81 \\
        MSL \cite{li2022MSL}            & AAAI22 & Uni & E & 78.28 \\
        UMIL \cite{lV2023UMIL}           & CVPR23 & Uni & E & 81.66 \\
        CU-Net \cite{zhang2023exploiting}         & CVPR23 & Uni & E & 78.74 \\
        TPWNG \cite{yang2024text}              & CVPR24 & Uni & E & 83.68\\
        LAVAD  \cite{zanella2024harnessing}            & CVPR24 & Uni & E & 62.01 \\
        VERA   \cite{ye2024vera}            & CVPR25 & Uni & E & 70.54 \\
        \rowcolor[HTML]{D4EDDA} DSRL* \cite{leng2024beyond}  & NeurIPS24 & Uni & E \& H & 82.01 \\
        \rowcolor[HTML]{C6E2A3} PiercingEye& - & Uni & E \& H & \textbf{83.74}\\
        \midrule
        HL-Net \cite{Wu2020HLnet}         & ECCV20  & Multi & E & 78.64 \\
        Wu et al. \cite{Wu2022weakly}      & TMM22   & Multi & E & 78.64 \\
        Pang et al. \cite{Pang2022AVD}    & TMM22   & Multi & E & 79.37 \\
        UMIL \cite{lV2023UMIL}          & CVPR23  & Multi & E & 81.77 \\
        CU-Net \cite{zhang2023exploiting}         & CVPR23  & Multi & E & 81.43 \\
        MACIL-SD \cite{Yu2022MACIL-SD}       & MM23    & Multi & E & 83.40 \\
        HyperVD \cite{HyperVD}        & IVC24   & Multi & H & 85.67 \\
        CFA-HLGAtt \cite{ghadiya2024cross}        & CVPRW24 & Multi & H & 86.34\\
        \rowcolor[HTML]{D4EDDA} DSRL* \cite{leng2024beyond} & NeurIPS24 & Multi & E \& H & 87.61\\
        \rowcolor[HTML]{C6E2A3} PiercingEye & - & Multi & E \& H & \textbf{88.82}\\
        \bottomrule
    \end{tabular}
    }
    \label{tab:XD_performance}
    \vspace{-12pt}
\end{table}
\begin{table}[t]
    \centering
    \scriptsize  
    \caption{Comparisons of frame-level AUC performance on UCF-Crime dataset under unimodal input setting. UCF-Crime only has visual modality input.``*" refers to our previous work.
}
    \label{tab:ucfcrimes}
    
    \setlength{\arrayrulewidth}{0.3mm}  
    \renewcommand{\arraystretch}{1.2}  
    \setlength{\tabcolsep}{6pt}  
    
    % 调整表格宽度
    \resizebox{\linewidth}{!}{
    \rowcolors{2}{gray!10}{white}  
    \begin{tabular}{l l c c c}  
        \toprule
        \rowcolor{black!10} \textbf{Methods} & \textbf{Venue} & \textbf{Input Setting} & \textbf{Feature Space} & \textbf{AUC(\%)}\\
        \midrule
        Sultani et al. \cite{Sultani} & CVPR18 & Uni & E & 76.21 \\
        Wu et al. \cite{Wu2021LCTR}   & TIP21  & Uni & E & 82.44 \\
        MIST \cite{Feng2021mist}  &  CVPR21& Uni& E&82.30\\
        RTFM \cite{Tian2021RTFM}              & ICCV21 & Uni & E & 84.30 \\
        MSL \cite{li2022MSL}                 & AAAI22 & Uni & E & 85.30 \\
        UMIL \cite{lV2023UMIL}              & CVPR23 & Uni & E & 86.75 \\
        CU-Net \cite{zhang2023exploiting} & CVPR23 & Uni & E & 86.22 \\
        UR-DMU \cite{zhou2023dual}        & AAAI23 & Uni & E & 86.97\\
        PEL4VAD  \cite{pu2024learning}    & TIP24  & Uni & E & 86.76\\
        LAVAD \cite{zanella2024harnessing}                        & CVPR24 & Uni & E & 80.28 \\
        VERA \cite{ye2024vera}                         & CVPR25 & Uni & E & 86.55 \\
        \rowcolor[HTML]{D4EDDA} DSRL* \cite{leng2024beyond}& NeurIPS24 & Uni & E \& H & 86.38 \\
        % PiercingEye 绿色背景
       \rowcolor[HTML]{C6E2A3} \textbf{PiercingEye} & - & Uni & E \& H & \textbf{86.64}\\
        
        \bottomrule
    \end{tabular}
    }
    \vspace{-10pt}
\end{table}
\begin{table}[t]
    \centering
    \scriptsize 
    \caption{Comparisons of frame-level AUC performance on a collected subset of UCF-Crime under unimodal input setting. UCF-Crime only has visual modality input.``*" refers to our previous work.
}
    \label{tab:subset}
    
    \setlength{\arrayrulewidth}{0.3mm}  
    \renewcommand{\arraystretch}{1.2}  
    \setlength{\tabcolsep}{6pt}

    \resizebox{\linewidth}{!}{
    \rowcolors{2}{gray!10}{white}  
    \begin{tabular}{l l c c cc}  
        \toprule
        \rowcolor{black!10} \textbf{Methods} & \textbf{Venue} & \textbf{Input Setting} & \textbf{Feature Space} & \textbf{UCF-Crime}&\textbf{Collected Subset}\\
        UR-DMU \cite{zhou2023dual} & AAAI23 & Uni & E & 86.97&80.88 \textcolor{blue}{ ($\downarrow$ 6.09)}\\
        PEL4VAD \cite{pu2024learning} & TIP24& Uni& E&  86.76&81.43 \textcolor{blue}{ ($\downarrow$ 5.33)}\\
        \rowcolor[HTML]{D4EDDA} DSRL* \cite{leng2024beyond}& NeurIPS24 & Uni & E \& H & 86.38&82.34 \textcolor{blue}{ ($\downarrow$ 4.04)}\\
        % PiercingEye 绿色背景
       \rowcolor[HTML]{C6E2A3} \textbf{PiercingEye} & - & Uni & E \& H & 86.64&\textbf{83.21} \textcolor{blue}{ ($\downarrow$ 3.43)}\\
        
        \bottomrule
    \end{tabular}
    }
    \vspace{-12pt}
\end{table}
\subsection{Comparisons with State-of-the-art Methods}
\begin{figure*}[ht]
    \centering
    \includegraphics[width=1.0\linewidth]{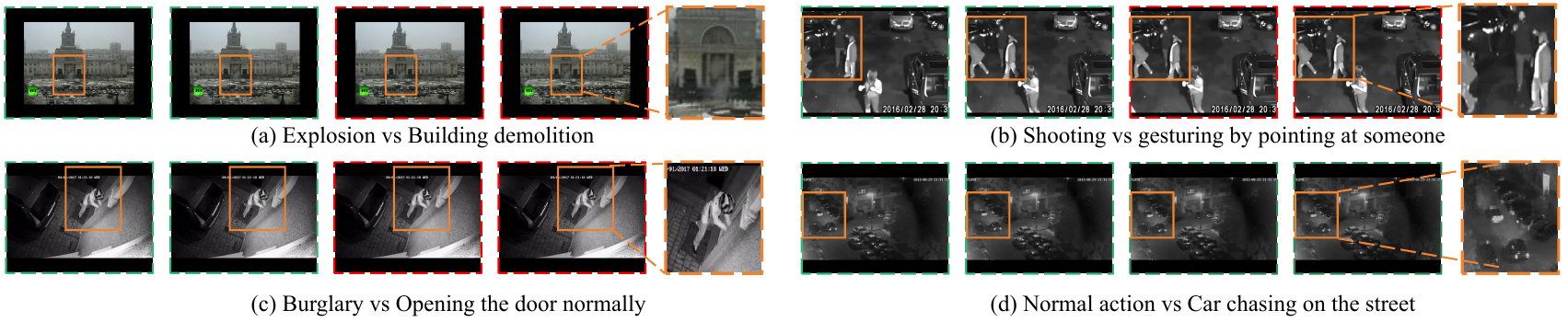} 
    \caption{Examples of a collected subet of UCF-Crime.     }
    \label{fig:vis_subset}
    \vspace{-16pt}
\end{figure*}

In this section, we evaluate the performance of PiercingEye on the XD-Violence dataset under both unimodal and multimodal input settings. We further assess the generalizability of the model by comparing it with previous state-of-the-art methods on the UCF-Crime dataset.\\
\textbf{Performance on XD-Violence under the multimodal setting.}
As shown in Table~\ref{tab:XD_performance}, PiercingEye achieves a notable performance gain under the multimodal input setting, surpassing the best Euclidean-based method by 5.42\% and the best hyperbolic-based method by 2.48\%. These results demonstrate the effectiveness of our dual-space representation in capturing complementary features from both geometries. Moreover, the superior performance over DSRL further highlights the contribution of the AETG and HVLGL modules to the overall feature learning process.\\ 
\textbf{Performance on XD-Violence under the unimodal setting.}
PiercingEye also sets a new state-of-the-art under the unimodal input setting, as shown in Table~\ref{tab:XD_performance}. Even with limited modality inputs, the model exhibits robust feature learning and maintains strong discriminative capability. While unimodal models often struggle with subtle semantic distinctions, our integration of Euclidean and hyperbolic representations enables PiercingEye to effectively capture fine-grained differences. \\
\textbf{Performance on UCF-Crime under the unimodal setting.}
To further validate the generalizability of our method, we evaluate it on the UCF-Crime dataset, which contains only visual modality. PiercingEye achieves a competitive AUC of 86.64\%, as reported in Table~\ref{tab:ucfcrimes}. This result is comparable with the current state-of-the-art methods, indicating that PiercingEye not only performs well on XD-Violence but also generalizes effectively to different datasets and domains.\\
\textbf{Evaluation on a Collected Subset of UCF-Crime.}
Since our method is primarily designed to address ambiguous events, and the UCF-Crime dataset contains relatively few such cases, the overall performance improvement on this dataset is limited. To further validate the effectiveness of our approach in handling ambiguous events, we designed an additional experiment. 
We first employed the PEL4VAD method \cite{pu2024learning} to evaluate all videos in the UCF-Crime test set and selected those with lower detection performance as candidate samples. To ensure these videos align with our definition of ``ambiguous events'', three annotators independently reviewed the low-performing samples. Their selections were then consolidated to construct a small-scale ambiguous event subset.
This subset contains 13 videos with a total of 32154 frames. We evaluated several representative methods on this dataset, including UR-DMU \cite{zhou2023dual}, PEL4VAD \cite{pu2024learning}, and DSRL \cite{leng2024beyond}, alongside our own method. As shown in Table \ref{tab:subset}, our method demonstrates clear advantages in handling ambiguous cases, further validating its effectiveness in recognizing visually similar but semantically diverse anomalies. 
To further confirm the reliability of this subset, we visualized several representative samples as shown in Fig. \ref{fig:vis_subset}. Most of them within the subset exhibit strong visual similarity while differing semantically, which is consistent with the characteristics of ambiguous events. 
This not only confirms the reliability of our subset construction but also highlights the superior performance of our method in handling such challenging ambiguous cases.
\subsection{Ablation Studies}
\subsubsection{Component-wise ablations}
To investigate the impact of each component in PiercingEye, including HE-GCN, DSI, and HVLGL, we begin with a baseline model that only utilizes GCN in Euclidean space and progressively add each component. Table \ref{tab:Ablations} presents the results on XD-Violence, starting with the baseline model, which shows relatively poor performance as it learns representations solely in Euclidean space ($1^{st}$ row), limiting its ability to capture hierarchical semantics. Similarly, learning only in hyperbolic space also yields suboptimal results ($2^{nd}$ row) due to weaker visual feature expressiveness. 
Next, we introduce HE-GCN, where the representations from different spaces are concatenated. This step benefits from the hierarchical context of events, leading to an improvement of 2.42\% in the multimodal setting and 1.75\% in the unimodal setting ($3^{rd}$ row).
Following this, we incorporate the DSI module to facilitate interactions between the different representations. This addition further boosts the performance, with an increase of 1.15\% in the multimodal setting and 2.31\% in the unimodal setting ($4^{th}$ row).
Finally, the HVLGL module is introduced, which generates ambiguous event text and leverages the loss function to guide the model in learning more discriminative features. This module results in notable performance improvements in both settings, achieving an increase of 1.21\% in the multimodal setting and 1.73\% in the unimodal setting ($5^{th}$ row). These results demonstrate the effectiveness of each individual component and highlight the overall improvement in performance as each module is added.
\begin{table}[t]
    \centering 
    \caption{Ablation studies of our method, measured by AP(\%) on XD-Violence dataset.}
    \resizebox{0.48\textwidth}{!}{
    \begin{tabular}{c c c c cc}
        \toprule
        \multirow{2}{*}{\textbf{Euclidean}}& \multirow{2}{*}{\textbf{Hyperbolic}} & \multirow{2}{*}{\textbf{DSI}} & \multirow{2}{*}{\textbf{HVLGL}} & \multicolumn{2}{c}{\textbf{XD-Violence (\%)}} \\
        \cmidrule(lr){5-6} 
        &  &  &  & \textbf{Unimodal} & \textbf{Multimodal} \\
        \midrule
        
        \checkmark & - & - & - & 77.95& 84.04\\
 -& \checkmark& -& -& 78.55&83.67\\
        \checkmark & \checkmark& -& -& 79.70& 86.46\\
        \checkmark & \checkmark & \checkmark& - & 82.01& 87.61\\
        \checkmark & \checkmark & \checkmark& \checkmark & \textbf{83.74}& \textbf{88.82}\\
        \bottomrule
    \end{tabular}
    }
    \label{tab:Ablations}
    \vspace{-10pt}
\end{table}
\begin{table}[t]
    \centering
    \caption{Ablation study of the proposed HE-GCN, measured by AP(\%) on XD-Violence dataset.}
    \label{tab:Ablations_HEGCN}
  \footnotesize
    \renewcommand{\arraystretch}{1.2}
    \resizebox{0.35\textwidth}{!}{
    \begin{tabular}{cc|cc}
        \toprule
        \multirow{2}{*}{\textbf{HGCN}}& \multirow{2}{*}{\textbf{HE-GCN}} & \multicolumn{2}{c}{\textbf{XD-Violence (AP\%)}} \\
        \cmidrule(lr){3-4}
        & & \textbf{Unimodal} & \textbf{Multimodal} \\
        \midrule
        \checkmark & --         & 80.33 & 84.41 \\
        --& \checkmark & \textbf{83.74} \textcolor{red}{($\uparrow$ 3.41)}& \textbf{88.82} \textcolor{red}{($\uparrow$ 4.41)}\\
        \bottomrule
    \end{tabular}
    }
    \vspace{-10pt}
\end{table}
\subsubsection{Evaluation of Different Message Passing Strategies}
\begin{figure}[t]
    \centering
    \includegraphics[width=1.0\linewidth]{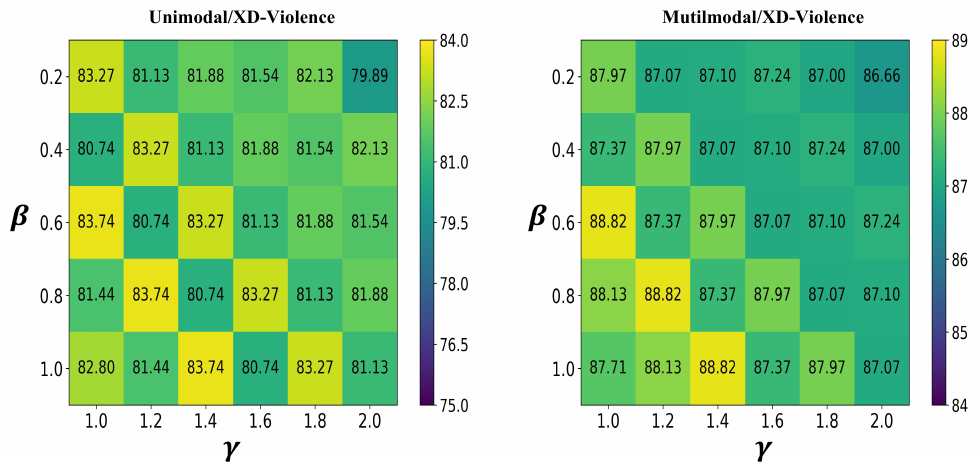} 
    \caption{Ablation syudies on $\beta$ and $\gamma$. We report the performance of the PirecingEye under different HE-GCN setups in both unimodal and mutilmodal seetings. Experiments are conducted on the XD-Violence dataset. 
    }
    \label{fig:hyper_beta}
    \vspace{-12pt}
\end{figure}

\textbf{The effectiveness of the LSHAD in HE-GCN.}  
In our method, we emphasize the necessity of constructing distinct message graphs at different layers to enable more effective message passing and to better capture hierarchical event context. LSHAD is a crucial element of our HE-GCN for constructing the message graph.
In Table \ref{tab:Ablations_HEGCN} ($2^{nd}$ and $3^{rd}$ rows), we present the results comparing HGCN used in HyperVD \cite{HyperVD}, which employs a hard node selection strategy (nodes selected by a fixed threshold), with HE-GCN, which uses our introduced layer-sensitive hyperbolic association degrees for node selection in message aggregation.
The performance of AP on XD-Violence improved by 4.41\% in the multimodal setting and by 3.41\% on the unimodal setting. These results demonstrate that our \textit{LSHAD} is a better strategy and helps to capture the hierarchical context of events.\\
\textbf{Reasons for the design choices in LSHAD with its multiple hyperparameters and threshold criteria.}
Inspired by the Global-first principle \cite{chen1982topological} that humans always have cognition on global first and then focus on local, we propose a novel node selection strategy, which guarantees the model captures the broader global context first with a relaxed threshold at the beginning of message aggregation and then focuses on the local context with more strict thresholds. 
To achieve this, we introduce the LSHAD construction rule, which calculates an LSHAD threshold based on the number of the current layer $K$ and hyperbolic Dirichlet energy of the current layer. As the $K$ increases and the hyperbolic Dirichlet energy decreases, the LSHAD threshold increases and is limited to 0 and 1 by the sigmoid function. If there is no $\beta$ and $\gamma$, the threshold in the first layer will be strict ($> 0.5$), causing the overlook of some global context information. Therefore, to make our node selection threshold conform to the Global-first principle, we introduced the two hyperparameters in LSHAD, where $\beta$ controls the influence of the number of current layer $ k$ and $\gamma$ acts as a bias to fine-tune the threshold. \\
\textbf{Sensitivity of Hyper-Parameters $\beta$ and $\gamma$.}
We employed a grid search to determine the optimal values of the two hyperparameters ($\beta$, $\gamma$), where $\beta$ ranges from [0.2, 0.4, 0.6, 0.8, 1.0] and $\gamma$ ranges from [1.0, 1.2, 1.4, 1.6, 1.8, 2.0]. As shown in Fig. \ref{fig:hyper_beta}, the results demonstrate that our model is relatively robust to changes in the hyperparameters within certain ranges. Additionally, the optimal performance is achieved when ($\gamma$-$\beta$) equals 0.4, with the pair (0.8, 1.2) selected from this range. Specifically, when $\beta$ = 0.8, variations in $\gamma$ within the tested range do not significantly affect the model’s performance. Similarly, when $\gamma$ = 1.2, changes in $\beta$ also have minimal impact on the model’s effectiveness. This suggests that our model’s performance remains stable under small perturbations of these hyperparameters, highlighting the robustness of the parameter configuration.

\subsubsection{Evaluation of Dual-Space Interaction Module}
\begin{figure}[t]
    \centering
    \includegraphics[width=1\linewidth]{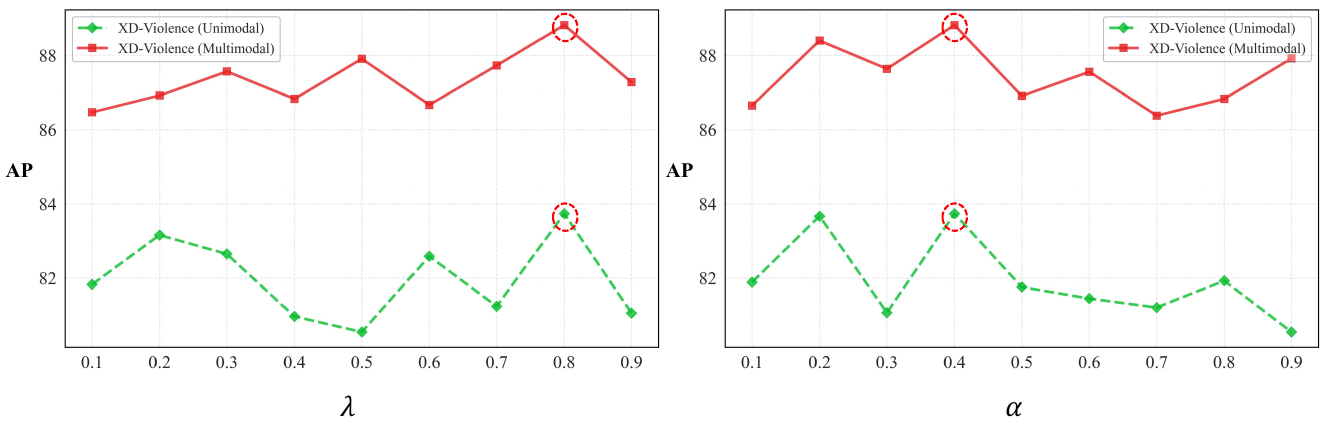} 
    \caption{Ablation studies on hyperparameters ($\lambda$ and $\alpha$) in the DSI module. 
    The left plot shows the performance (AP) of the model with varying values of $\lambda$ for both XD-Violence (unimodal) and XD-Violence (multimodal) settings. The red circle highlights the optimal value of $\lambda = 0.8$, where the model achieves the best performance. The right plot illustrates the performance with varying values of $\alpha$, and the red circle marks the optimal value of $\alpha = 0.4$, where the best results are obtained. These observations suggest that both hyperparameters are critical to improving model performance across both unimodal and multimodal settings.}
    \label{fig:hyper_alpha}
    \vspace{-10pt}
\end{figure}

\textbf{Lorentzian metric in DSI.} The DSI module is designed to improve information interaction between features from two different geometric spaces. To evaluate its effectiveness, we compared the impact of cosine similarity and the Lorentzian metric. As shown in Table \ref{tab:Ablations_DSI}, using cosine similarity for attention calculation results in a performance drop of 2.03\% in the unimodal setting and 2.01\% in the multimodal setting compared to the Lorentzian metric. This suggests that the Lorentzian metric is more effective in capturing feature similarities across different spaces, improving information interaction.
The main advantage of the Lorentzian metric is its ability to compute nonlinear distances between nodes, which better captures the complex structure of data. This is particularly useful for tasks like violence detection, where relationships between events are nonlinear. In contrast, cosine similarity calculates linear distances, which can lead to false relationships. This is problematic in ambiguous violence events, such as distinguishing sports collisions from violent actions, where visual cues are similar but semantic meanings differ. Cosine similarity struggles to differentiate these, making it less effective in such cases.
By using the Lorentzian metric, our model gains a more accurate measure of feature similarity, improving its ability to distinguish visually similar but contextually different events. This enhances the overall performance of the DSI module and highlights the importance of choosing the right distance metric for cross-space interaction.\\ 
\textbf{Ablation studies on Hyper-Parameters $\alpha$ and $\lambda$.} We conduct ablation studies on several hyperparameters in the DSI module, and the experimental results are presented in Fig. \ref{fig:hyper_alpha}. In Eq. \ref{EGA}, $\lambda$ is the threshold value used to eliminate weak relationships while strengthening the correlations between more similar node pairs. The primary goal of this thresholding mechanism is to reduce the influence of irrelevant or insignificant nodes, allowing the model to focus only on node pairs with strong similarity. This enhances the model's attention mechanism, enabling it to concentrate on the most important relationships. As shown in Fig. \ref{fig:hyper_alpha}, we observe that when $\lambda$ is set to 0.8, the best performance is achieved in both unimodal and multimodal settings.
In Eq. \ref{CSA}, $\alpha$ serves as a scaling factor that controls the contribution of hyperbolic space features ($V_H$) to the enhanced Euclidean space features ($V_E^\prime$). Specifically, $\alpha$ adjusts the influence of the cross-space attention mechanism ($CSA_{E \to H}$) in modifying the final enhanced features. As observed in Fig. \ref{fig:hyper_alpha}, the best performance is achieved when $\alpha$ is set to 0.4. This indicates that $\alpha$ plays a crucial role in balancing the contributions of both spaces, optimizing the overall performance of the model.
\begin{table}[t]
    \centering
    \caption{Ablation studies of the proposed DSI, measured by AP(\%) on XD-Violence dataset.}
    \label{tab:Ablations_DSI}
    \footnotesize
    \renewcommand{\arraystretch}{1.2}
    \resizebox{0.5\textwidth}{!}{
    \begin{tabular}{ccc|cc}
        \toprule
        \multirow{2}{*}{\textbf{Concat}}& \multirow{2}{*}{\textbf{Cosine Metric}} &\multirow{2}{*}{\textbf{Lorentzian metric}}& \multicolumn{2}{c}{\textbf{XD-Violence (AP\%)}} \\
        \cmidrule(lr){4-5}
        &  & & \textbf{Unimodal} & \textbf{Multimodal} \\
        \midrule
        \checkmark & --          &--& 81.71& 86.81\\
        --& \checkmark  &--&80.50& 86.89\\
 --& --& \checkmark& \textbf{83.74} \textcolor{red}{($\uparrow$ 2.03)}&\textbf{88.82} \textcolor{red}{($\uparrow$ 2.01)}\\
 \bottomrule
    \end{tabular}
    }
\end{table}
\begin{table}[t]
    \centering
    \caption{Ablation studies of the proposed HVLGL, measured by AP(\%) on XD-Violence dataset.}
    \label{tab:VL-loss}
    \footnotesize
    \renewcommand{\arraystretch}{1.2}
    \resizebox{0.5\textwidth}{!}{
    \begin{tabular}{ccc|cc}
        \toprule
        \multirow{2}{*}{\textbf{Euclidean-Loss}}& \multirow{2}{*}{\textbf{Hyperbolic-Loss}}&\multirow{2}{*}{\textbf{HVLGL}}& \multicolumn{2}{c}{\textbf{XD-Violence (AP\%)}} \\
        \cmidrule(lr){4-5}
        &  & & \textbf{Unimodal} & \textbf{Multimodal} \\
        \midrule
        \checkmark & --          &--& 79.43 & 87.51 \\
        --& \checkmark  &--&82.91& 87.66\\
 --& --& \checkmark& \textbf{83.74} \textcolor{red}{($\uparrow$ 4.31)}&\textbf{88.82} \textcolor{red}{($\uparrow$ 1.16)}\\
 \bottomrule
    \end{tabular}
    }
    \vspace{-8pt}
\end{table}
\begin{table}[t]
    \centering
    \caption{Ablation studies on hyperparameters $\tau$ and $\theta$, measured by AP (\%) in XD-Violence dataset under multimodal setting.}
    \label{tab:tau_theta_ablation}
    \renewcommand{\arraystretch}{1.2}
    \rowcolors{2}{gray!10}{white}
    \resizebox{0.48\textwidth}{!}{
    \begin{tabular}{c|cc|cc|cc}
        \toprule
        \multirow{2}{*}{$\tau$} & \multicolumn{2}{c|}{$\theta = 1.0$} & \multicolumn{2}{c|}{$\theta = 2.0$} & \multicolumn{2}{c}{$\theta = 3.0$} \\
        \cmidrule(lr){2-3} \cmidrule(lr){4-5} \cmidrule(lr){6-7}
        & Multimodal& Unimodal& Multimodal& Unimodal& Multimodal& Unimodal\\
        \midrule
        0.03 & 85.79 & 80.29 & 86.61 & 81.84 & 86.55 & 81.58 \\
        0.05 & 87.11 & 81.44 & 86.75 & 81.11 & 85.32 & 80.97 \\
        0.07 & 87.01 & 81.21 & 86.24 & 81.59 & 86.73 & 80.08 \\
        0.10 & 86.84 & 80.76 & 86.78 & 82.60& 87.33& 80.72 \\
        0.30 & \textbf{88.82} & \textbf{83.74} & 88.42& 82.11 & 87.74 & 81.26\\
        0.50 & 86.93 & 80.84 & 86.03 & 83.59 & 86.48 & 83.04 \\
        0.70 & 87.25 & 82.43 & 86.41 & 80.35 & 86.84 & 81.24 \\
        \bottomrule
    \end{tabular}
    }
    \vspace{-12pt}
\end{table}
\begin{figure}[t]
    \centering
    \includegraphics[width=0.9\linewidth]{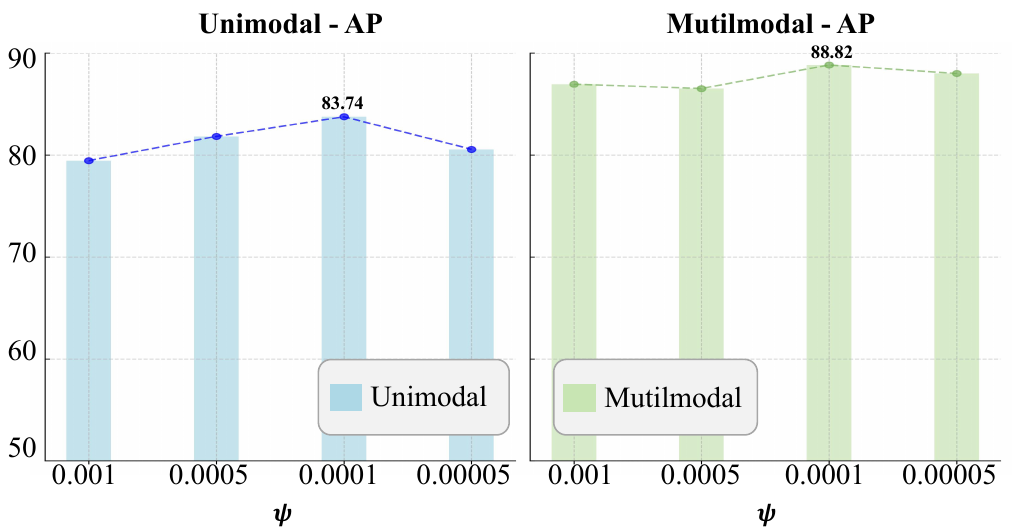} 
    \caption{Performance comparison of XD-Violence dataset under unimodal and multimodal settings with different values of \(\psi\) for the HVLGL loss weight. The values represent the Average Precision (AP) scores. The optimal performance is observed at \(\psi = 0.0001\), showing a significant improvement in the multimodal task. When \(\psi\) becomes too large, the performance drops, indicating that excessively high \(\psi\) values might negatively affect the model's learning.
    }
    \label{fig:hyper_psi}
    \vspace{-16pt}
\end{figure}
\subsubsection{Evaluation of HVLGL module}
\textbf{Comparison of Different Vision-Text Contrastive Losses.} 
In our approach, we focus on leveraging generated ambiguous event texts to guide the model's learning. Here, we compare three types of contrastive losses: (1) InfoNCE loss \cite{oord2018representation}, the commonly used Euclidean contrastive loss (denoted as``Euclidean-Loss"); (2) the hyperbolic contrastive loss (denoted as ``Hyperbolic-Loss"), which uses a hyperbolic distance metric to measure similarity between samples; and (3) the proposed HVLGL, which incorporates text-based similarity into the hyperbolic contrastive loss to dynamically weight negative samples.
As shown in the second and third rows of Table \ref{tab:VL-loss}, using Hyperbolic-Loss outperforms Euclidean-Loss by 3.48\% and 0.15\% under unimodal and multimodal settings, respectively. This demonstrates the superiority of the Lorentzian metric in capturing relationships between samples compared to the Euclidean metric, also reflected in Table \ref{tab:Ablations_DSI}. Furthermore, comparing the third and fourth rows, HVLGL brings an additional improvement of 0.83\% and 1.16\% in the unimodal and multimodal settings, respectively. This indicates that incorporating text similarity to measure the difficulty of negative samples and dynamically control their separation is an effective strategy for enhancing contrastive learning.\\
\textbf{Analysis of Hyper-Parameter $\tau$, $\theta$ and $\psi$.}
The ablation study results in Table \ref{tab:tau_theta_ablation} show that both $\tau$ and $\theta$ significantly impact the model's performance. In both unimodal (83.74\%) and multimodal (88.82\%) settings, the model achieves the best performance when $\tau = 0.3$. This indicates that a larger $\tau$ helps the model more effectively distinguish between positive and negative samples, enhancing its ability to capture relationships between complex samples. However, when $\tau$ is too large (e.g., greater than 0.3), the model may overly focus on extreme sample pairs and neglect finer sample differences, leading to a decrease in performance. Therefore, the value of $\tau$ should be moderate; a larger $\tau$ can improve discriminative ability, but if it becomes too large, it may cause overfitting, reducing the model’s generalization ability. On the other hand, when $\tau$ is smaller (e.g., 0.03 or 0.05), performance drops, especially in the unimodal setting, suggesting that smaller $\tau$ values cause the model to focus too much on simple sample pairs, thus weakening its discriminative power.

For $\theta$, we observe that increasing $\theta$ generally enhances performance, especially in the unimodal setting. Larger values of $\theta$ (e.g., $\theta = 3.0$) help the model focus more on harder negative samples, which is beneficial for improving its performance on challenging tasks. However, for the multimodal setting, performance starts to slightly degrade with very high $\theta$ values (e.g., $\theta = 3.0$), likely because the model overemphasizes difficult negatives and neglects other important sample relationships. Thus, both $\tau$ and $\theta$ should be balanced to achieve optimal performance without overfitting to difficult samples.

In the experiment, \(\psi\), as the weight of the HVLGL loss, adjusts the model's focus on the HVLGL loss versus the classification loss. Fig. \ref{fig:hyper_psi} shows that when \(\psi = 0.0001\), the model performs the best in both unimodal and multimodal settings, with the AP value reaching 88.82\% in the multimodal task. However, when \(\psi\) is too high (e.g., \(\psi = 5e-5\)), the model's performance decreases. This indicates that an overly high \(\psi\) value overly emphasizes the HVLGL loss, causing the model to neglect the balance of dual-space feature learning, which affects the feature fusion effectiveness. Therefore, a reasonable \(\psi\) value helps to find a balance between the classification loss and the HVLGL loss, improving the model's performance in multimodal tasks. 

\subsection{Qualitative Analysis}
\subsubsection{Qualitative Visualizations of PiercingEye in the Context of Ambiguous Violence}
\begin{figure*}[ht]
    \centering
    \includegraphics[width=1\linewidth]{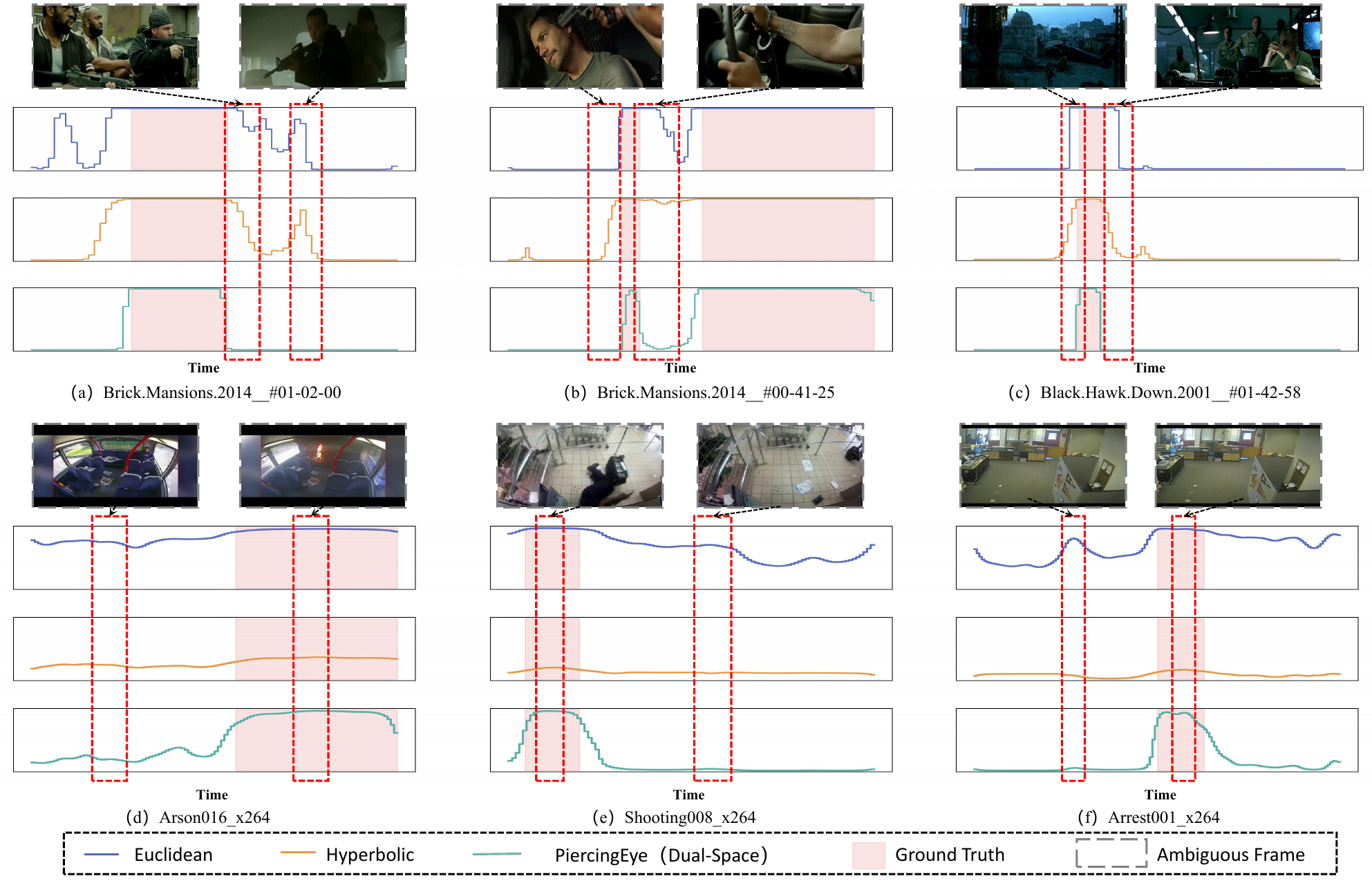} 
    \caption{Some visual results of PiercingEye in the context of ambiguous violence. The blue curves show violence scores predicted using only Euclidean representation. The yellow curve shows scores using only hyperbolic representation (It is worth noting that on the XD-Violence dataset, we use HyperVD as the representative method that employs only hyperbolic representations, while on the UCF-Crime dataset, we evaluate using our own model trained with HEGCN alone.). The green curves show scores predicted by PiercingEye, and the pink area represents the ground-truth violent temporal location. (a)–(c) are from the XD-Violence dataset, while (d)–(f) are from the UCF-Crime dataset.
    }
    \label{fig:amibigious_vis}
    \vspace{-16pt}
\end{figure*}

We put up qualitative visualizations of PiercingEye when handling ambiguous events. Fig. \ref{fig:amibigious_vis} demonstrates that PiercingEye effectively resolves ambiguous events, with which single-space representation learning struggles. 

In Fig. \ref{fig:amibigious_vis}, the first row is from the XD-Violence dataset, and the second row is from the UCF-Crime dataset. Taking example (a) in the Fig. \ref{fig:amibigious_vis} , the first ambiguous frame shows several men holding guns, which appears visually abnormal. The Euclidean space representation, relying on local visual cues, immediately classifies this frame as anomalous. The hyperbolic space representation, influenced by the preceding violent context, also predicts it as an anomaly. However, PiercingEye effectively integrates both perspectives: while the Euclidean space identifies the presence of guns as a potential threat, the hyperbolic space captures the broader context, suggesting that the situation is actually under control. Through this fusion of information, PiercingEye is able to accurately classify the frame as non-anomalous. 
This supports our motivation: hyperbolic representation enhances hierarchical event relations but weakens visual feature expression, while Euclidean representation emphasizes visual features but overlooks event relationships. PiercingEye effectively addresses ambiguous violence, which is challenging for either space alone.
\subsubsection{Feature Discrimination Visualization}
\begin{figure*}[ht]
    \centering
    \includegraphics[width=1\linewidth]{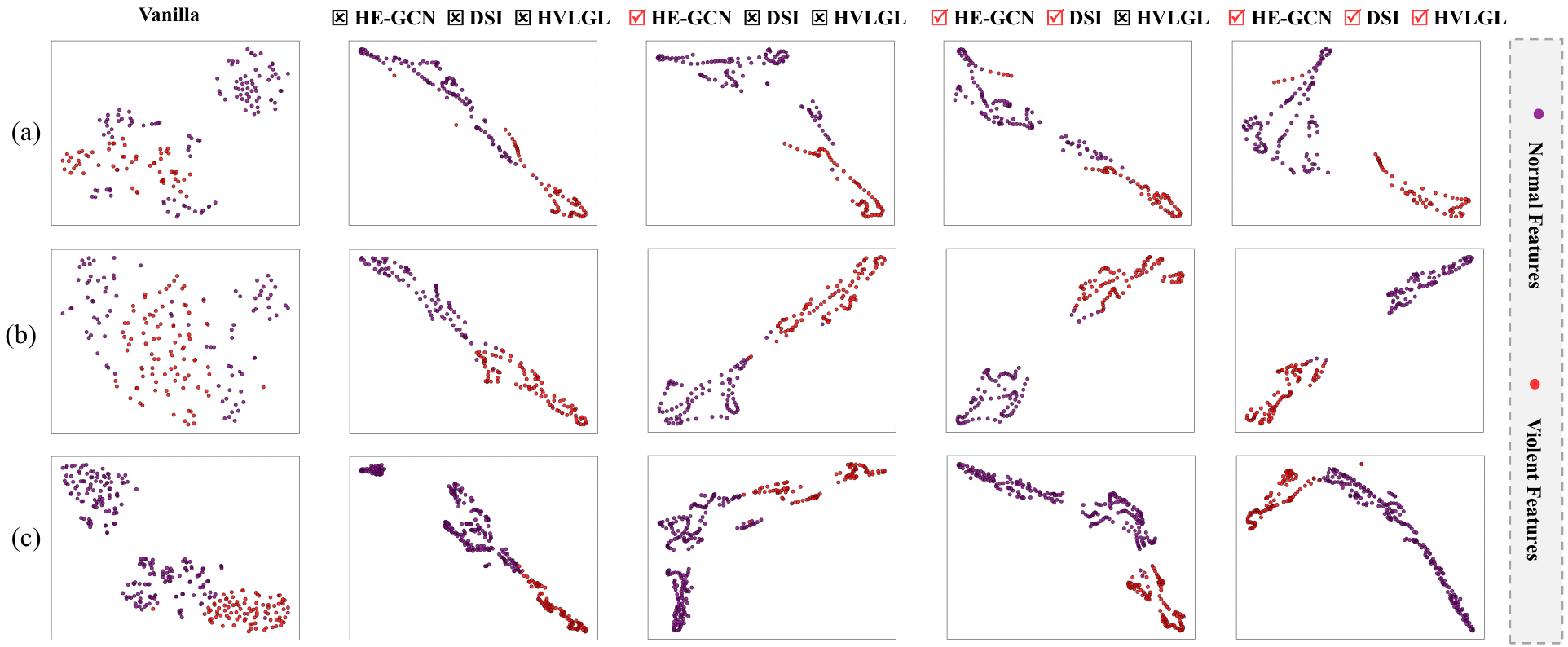} 
    \caption{t-SNE visualization of feature distributions for normal (purple) and violent (red) samples under different ablation settings. (a)-(c) are from XD-Violence test set. The five configurations are compared from left to right: Vanilla, Baseline (GCN), GCN + HE-GCN, GCN + HE-GCN + DSI, and the full PiercingEye model (GCN + HE-GCN + DSI + HVLGL). This visualization helps to understand the contribution of each module in enhancing feature separability and improving the model's ability to distinguish between normal and violent events.
    }
    \label{fig:Feature_dis}
    \vspace{-10pt}
\end{figure*}

To better understand the contribution of each module in our proposed PiercingEye framework, we conduct a t-SNE visualization of the extracted features under different ablation settings. Fig. \ref{fig:Feature_dis} shows the feature distributions of normal (purple) and violent (red) samples, where row (a) illustrates a representative example from the XD-Violence dataset. From left to right, we compare the feature separability across five configurations: Vanilla, Baseline (GCN), GCN + HE-GCN, GCN + HE-GCN + DSI, and the full model PiercingEye (GCN + HE-GCN + DSI + HVLGL).
\textbf{Vanilla:} Without any structural modeling, the vanilla model shows a clear overlap between violent and normal samples, indicating poor discriminability. This reflects the limitation of using basic representations without spatial or semantic guidance, making it particularly ineffective for distinguishing ambiguous events.
\textbf{Baseline (GCN):} Introducing GCN as the baseline enhances local relational modeling of visual features in Euclidean space. As a result, feature clusters become more structured, and the separation between normal and violent samples improves. However, some overlap remains, as Euclidean representations primarily focus on motion and appearance cues but struggle to capture global or hierarchical event semantics—limiting their ability to handle ambiguity.
\textbf{GCN + HE-GCN: }Adding the HE-GCN module brings hyperbolic geometry into play, allowing the model to encode global contextual and hierarchical relationships. The dynamic node selection mechanism, guided by layer-sensitive hyperbolic association degrees and Dirichlet energy, enables the model to better identify salient contextual patterns. Consequently, violent samples begin to form more compact and distinguishable clusters, validating the effectiveness of hyperbolic modeling in representing event hierarchies.
\textbf{GCN + HE-GCN + DSI:} Incorporating the Dual-Space Interaction (DSI) module significantly improves feature separability. DSI breaks the information isolation between Euclidean and hyperbolic spaces through cross-space attention, enabling complementary interactions: Euclidean space contributes strong visual discrimination, while hyperbolic space facilitates hierarchical event reasoning. This synergy is especially beneficial for modeling ambiguous scenarios, where relying on a single geometry often proves inadequate.
\textbf{PiercingEye (Full Model): }The final column shows the full model with all modules, including the Hyperbolic Vision-Language Guided Loss (HVLGL). Feature clusters become more semantically aligned and compact, while ambiguous samples are better positioned under the guidance of vision-language alignment. HVLGL introduces fine-grained supervision in hyperbolic space by aligning visual features with generated ambiguous textual descriptions, encouraging the model to focus on visually similar but semantically divergent instances. This, combined with the Ambiguous Event Text Generation (AETG) module—which systematically augments the training set with synthesized ambiguous event texts—further strengthens the model's ability to learn from scarce and hard-to-label ambiguous data.
\subsubsection{Visualization of AETG-Generated Ambiguous Texts}
\begin{figure*}[t]
    \centering
    \includegraphics[width=1\linewidth]{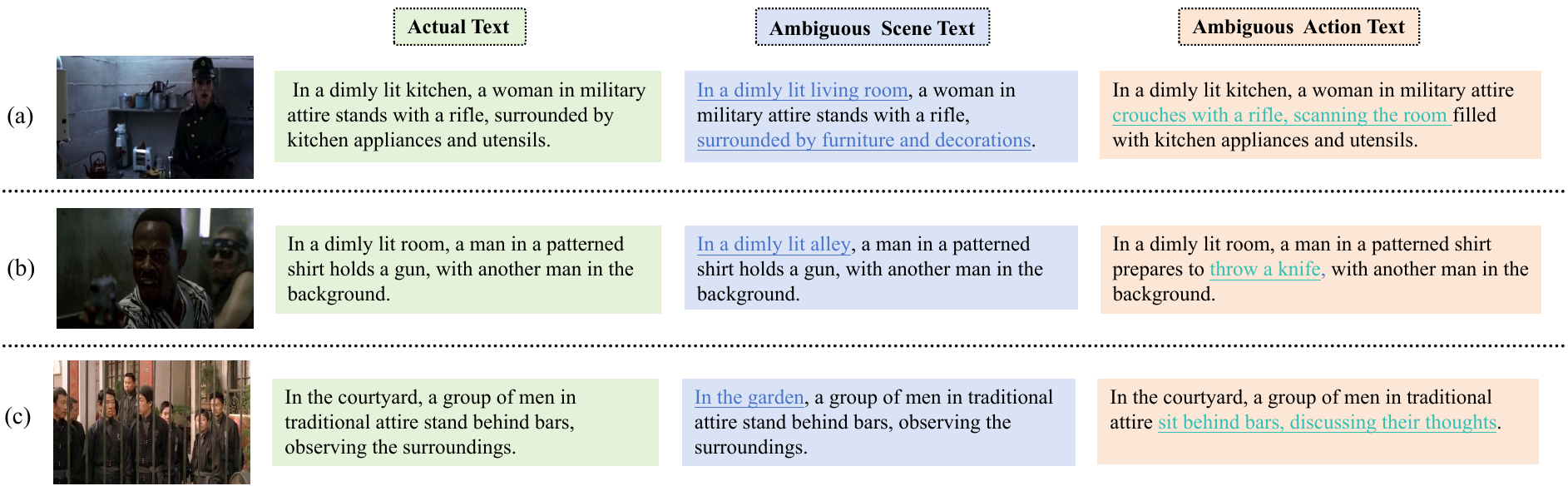} 
    \caption{Visualization of Ambiguous Texts Generated by AETG. (a)-(c) are from XD-Violence dataset.
    }
    \label{fig:AETG_text_vis}
    \vspace{-10pt}
\end{figure*}

To qualitatively evaluate the effectiveness of the Ambiguous Event Text Generation (AETG) module, we visualize examples of generated ambiguous descriptions in Fig.~\ref{fig:AETG_text_vis}. For each input video frame, we show the original ground-truth description (Actual Text), as well as two types of AETG-generated ambiguous texts: one by modifying the scene context (Ambiguous Scene Text) and the other by modifying the action semantics (Ambiguous Action Text). These examples demonstrate how AETG introduces controlled semantic ambiguity while preserving visual plausibility.
In example (a), the original caption describes``a woman in military attire stands with a rifle in a dimly lit kitchen''. The scene-ambiguous version replaces the kitchen with ``a dimly lit living room'', accompanied by ``furniture and decorations'', thereby shifting the perceived environment while maintaining consistency with the image. The action-ambiguous version instead modifies the behavior: the woman now ``crouches with a rifle, scanning the room'', introducing an interpretation that may suggest alertness or defensiveness rather than immediate aggression. This showcases AETG’s ability to generate semantically distinct but visually plausible descriptions, which provide valuable supervision for learning fine-grained and ambiguous visual concepts.
\subsubsection{Visualization of the Effectiveness of PiercingEye}
\begin{figure*}[ht]
    \centering
    \includegraphics[width=1\linewidth]{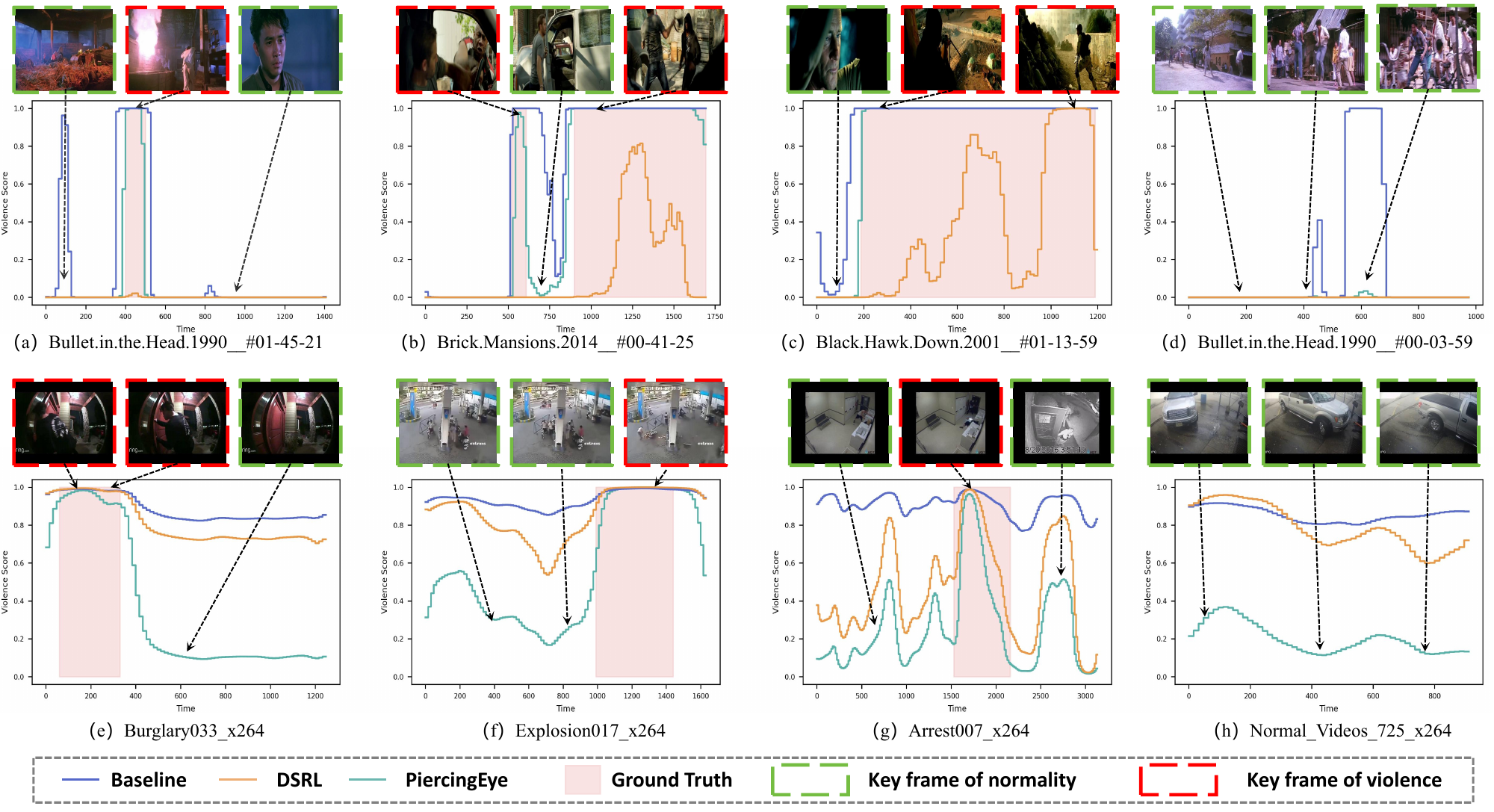}
    \caption{Frame-level scores and violence localization examples for the test video from XD-Violence dataset. 
(a)–(c) and (d)–(f) represent violent videos from the XD-Violence and UCF-Crime datasets, respectively. (g) depicts normal videos from XD-Violence, while (h)–(i) are from UCF-Crime. From the visualized curves, it is evident that our model demonstrates strong detection performance, especially when compared to the baseline and DSRL methods.
    }
    \label{fig:visualization}
    \vspace{-12pt}
\end{figure*}

We illustrate the qualitative visualizations of VVD for the test video from the XD-Violence dataset and UCF-Crime dataset. Fig. \ref{fig:visualization} shows that PiercingEye can accurately detect violent events and has a better detection performance than the baseline and DSRL \cite{leng2024beyond}. 
\subsection{Model Complexity and Speed }
Compared to DSRL \cite{leng2024beyond}, our AETG and HVLGL modules are used solely during training, adding virtually no additional parameters. The computational efficiency of the PiercingEye model is crucial, particularly for real-time applications. Our analysis confirms that the model meets real-time processing requirements, as evidenced by the following results. 
Our experiments were conducted on a single NVIDIA RTX A6000 GPU. For \textbf{video input}, the model processes at a rate of 83.87 FPS, handling only video data. The model’s parameters are relatively lightweight, totaling 13.4 MB, with I3D \cite{I3D} parameters at 12.49 MB and PiercingEye  parameters at 0.91 MB. This compact size ensures quick response times. 
When processing \textbf{video and audio inputs}, the model maintains a high processing speed of 56.86 FPS, despite the added computational load from audio processing. The total parameter size for this configuration is 85.54 MB, comprising I3D \cite{I3D} parameters at 12.49 MB, VGGish \cite{VGGish} parameters at 72.14 MB, and PiercingEye  parameters at 0.91 MB. 
These results illustrate that our model exhibits excellent real-time performance, even with multimodal inputs, making it suitable for latency-sensitive real-world applications.

\section{Conclusion}
\label{conclusion}
In this paper, we propose a comprehensive geometric representation learning method, PiercingEye, which integrates the benefits of Euclidean and hyperbolic geometries to improve the discrimination of ambiguous events. 
The hyperbolic energy-constrained graph convolutional network (HE-GCN) is designed to better capture the hierarchical context of events. Additionally, Dual-Space Interaction (DSI) is designed to facilitate information interactions. 
To explicitly enhance the model's discriminative ability, we propose the Ambiguous Event Text Generation (AETG) module, which leverages large language models to generate ambiguous text descriptions. We also design a novel Hyperbolic Vision-Language Guided Loss (HVLGL) that combines hyperbolic space with a dynamic weighting mechanism based on text similarity, ensuring that the model prioritizes learning from the most challenging ambiguous events, thereby improving the model's learning and generalization.
Our method achieves state-of-the-art performance on the XD-Violence dataset in both unimodal and multimodal settings, particularly excelling in resolving ambiguous events. Additionally, testing on a curated ambiguous event subset from the UCF-Crime dataset demonstrates the robustness and generalization of our approach in handling complex, ambiguous events. These results highlight PiercingEye's effectiveness in enhancing the accuracy and robustness of VVD in challenging scenarios. \\
\textbf{Future work.} Current datasets contain limited samples of ambiguous events, which constrains the generalization ability of existing models. In future work, we aim to address this limitation by expanding ambiguous event datasets with more diverse, fine-grained, and semantically rich samples across various scenarios. Additionally, we plan to further exploit the powerful semantic understanding of large vision-language models and large language models (VLMs/LLMs) to better generate, interpret, and align ambiguous event descriptions, ultimately enhancing the model’s capacity to handle complex and semantically confusing violent behaviors.

\bibliography{main}
\newpage

\vspace{11pt}

\vfill

\end{document}